\documentclass{article} 
\usepackage[margin=1.6in]{geometry}
\usepackage{graphicx}

\begin{document}

\title{Effects of Faults, Experience, and Personality on Trust in a Robot Co-Worker}
\author{Satragni Sarkar, Dejanira Araiza-Illan, and Kerstin Eder\footnote{Department of Computer Science, University of Bristol, UK, Emails: ss14379.2014@my.bristol.ac.uk, dejanira.araizaillan@bristol.ac.uk, kerstin.eder@bristol.ac.uk}}
\date{}

\maketitle

\begin{abstract}
To design trustworthy robots, we need to understand the impact factors of trust: people's attitudes, experience, and characteristics; the robot's physical design, reliability, and performance; a task's specification and the circumstances under which it is to be performed, e.g.\ at leisure or under time pressure.
As robots are used for a wide variety of tasks and applications, robot designers ought to be provided with evidence and guidance, to inform their decisions to achieve safe, trustworthy and efficient human-robot interactions. 
In this work, the impact factors of trust in a collaborative manufacturing scenario are studied by conducting an experiment with a real robot and participants where a physical object was assembled and then disassembled. 
Objective and subjective measures were employed to evaluate the development of trust, under faulty and non-faulty robot conditions, and the effect of previous experience with robots, and personality traits. 
Our findings highlight differences when compared to other, more social, scenarios with robotic assistants (such as a home care assistant), in that the condition (faulty or not) does not have a significant impact on the human's perception of the robot in terms of human-likeliness, likeability, trustworthiness, and even competence. However, personality and previous experience do have an effect on how the robot is perceived by participants, even though that is relatively small.

\end{abstract}

\section{Introduction}

The adoption of robotic assistants for industrial, military, research, medical, and service purposes is increasing. 
In an industrial setting, robotic co-workers such as Baxter\footnote{\label{fn:baxter}http://www.rethinkrobotics.com/baxter}, Sawyer\footnote{http://www.rethinkrobotics.com/sawyer-intera-3/} and Yumi\footnote{http://new.abb.com/products/robotics/yumi}, have been developed to support manufacturing tasks in close proximity to people. 
Manufacturing robotic assistants provide accuracy and reliability for high-quality production, redundancy to support the tasks of the human co-workers, and flexibility through reprogramming, customization, or even learning by example~\cite{Kruger2009,Hagele2012,Eder2014}. 

Historically, all industrial robots worked in isolation from people in limited and protected workspaces, for safety reasons~\cite{Eder2014}. 
The demands of flexible manufacturing now require robotic assistants to share a
workspace, and to directly interact with people through interfaces and diverse
communication means (e.g.\ voice commands) as well as joint manipulation of
objects. 
It is critical that system designers understand how trustworthiness in this kind of interaction can be established. 
While robotic designers can equip robots with measures that ensure people's safety~\cite{Desantis2008,Haddadin2011}, trust goes well beyond this, requiring people's perceptions to be understood for them to be considered in the design process.
For instance, trust can affect the performance of a worker involved in a
particular collaborative task, as workers would need to exchange information
with the robot, and perhaps even follow suggestions and directions~\cite{Salem2015} given by their robotic co-worker.
Hence, achieving high levels of trust in robotic co-workers is the basis for their acceptance, which in turn will open the doors for their widespread use and adoption on the factory floor.

To design trustworthy robots, we need to understand the impact factors of trust: how do people react to and perceive a robotic assistant? What are the human traits that cause this perception? Do people trust a robot that could potentially make mistakes during operation? Are ``friendly'' robots more trustworthy?
Trustworthiness factors, and the impact of faultiness in robots on people's trust in them, in  settings that differ wrt.\ safety, dependability, time pressure, as well as delivery, quality and performance demands need to be further investigated~\cite{Robinette2015} beyond social and basic assistance scenarios such as a museum guide~\cite{Burgard1998}, 
care of vulnerable people~\cite{Salem2015challenges} and 
search and rescue tasks~\cite{Desai2012,Desai2013}.
In the case of a collaborative industrial setting, would people feel comfortable and eased out, or stressed and threatened by their robotic co-workers?
If a robotic co-worker makes mistakes in an assembly line, how would this affect trust?

In this work we studied impact factors of trust in a collaborative manufacturing scenario, by conducting an experiment with a real robot and participants where a physical object was  assembled and then disassembled. 
Our experimental setup draws inspiration from a home care scenario, in which the role of errors by a faulty robot in establishing human--robot trust was studied, along with the influence of the type of task in the participants' willingness to follow a robot's instructions, and the effect of human personality traits~\cite{Salem2015,Salem2015challenges}.
The primary contribution of our work to human--robot interactions is deeper understanding of the impact of different kinds of faults, and participants' previous experience and personality traits, on trust within a collaborative manufacturing context.

Understanding what makes robotic assistants likeable and trustworthy allows designing better robotic co-workers, and thus maximises the potential of adopting these systems in a variety of applications. 
Our study demonstrates that cooperative manufacturing robots, flexible and low cost, but at the same time not 100\% accurate, could be well accepted in reality, as they were found likeable, safe, and trustworthy by the majority of participants. 
Furthermore, errors did not trigger a radical change in the participants' subjective perception of the robot.

The rest of this article is organized as follows. 
Section~\ref{sc:relatedwork} describes related work on measuring trust in human--robot interactions and robotic co-workers in particular. 
Section~\ref{sc:method} presents our hypotheses and the experimental design for the collaborative assembly scenario, in which participants assembled and disassembled an object, with guidance given by a faulty or non-faulty robot. 
Section~\ref{sc:results} reports on the results of our study. First, previous experience with technology and trust in the robot are evaluated before the experiment. Second, trust after the experiment is evaluated, along with personality traits, and the perceived robot's anthropomorphism and likeability. 
Section~\ref{sc:discussion} comments on the results and discusses our findings in the context of our hypotheses. 
Section~\ref{sc:conclusion} summarizes the article and presents our conclusions.

\section{RELATED WORK}\label{sc:relatedwork}

As robots make the leap from confined spaces and laboratories to our everyday lives, we are faced with interactions between people and physical technology that are richer than other computer-based systems or computerized characters (e.g.\ websites, online shopping, games, tablets), which have limited cognitive resources~\cite{Grabner2003,Kidd2004,Salem2015}. 
Human-robot interactions involve communication between a user and a (computer-based intelligent) mobile, physical system: the robot~\cite{Huttenrauch2006}. 
As robots are perceived differently (and potentially developing more complex relationships) than purely online systems, the study of trust development is paramount, and, although some research has been done within the field of experimental human-robot interactions, it needs to be extended to cover a substantial variety of potential applications, from `relaxed' social service scenarios to high-performance industrial settings.

\subsection{Measuring Trust in Human-Robot Interactions} 

The study of trust in the field of human-computer interactions is relevant for human-robot ones, as it provides insight in metrics, experimental methodologies, and causal factors.  
Although the study and measurement of trust varies in levels of agreement and effort, depending on the field of study (from philosophy to electronic commerce), research agrees on the value and importance of trust, as it enables living under risk and uncertainty, facilitates decisions, and influences cooperation, coordination, and long-term relationships~\cite{Corritore2003,Grabner2003}. 
Also, research on human-computer interactions has discovered that people enter into relationships with computers, websites and media, where social rules are applied and thus online systems are social actors. 
For these systems, trust has been defined as ``an attitude of confident expectation $[\ldots]$ that one's vulnerabilities will not be exploited''~\cite[p. 740]{Corritore2003}, ``willingness of a party to be vulnerable to the actions of another party''~\cite[p. 788]{Grabner2003}, or even ``the extent to which the [machine] is perceived to perform its function properly''~\cite[p. 442]{Muir1996}, which can be transferred to robots e.g.\ in terms of perceived task-specific safety and competence. 
Nonetheless, the specific external factors (environment, interfaces) and perception that influence trust assessment in human-computer interactions differ from the ones in human-robot interactions, due to the robot's physical embodiment.

In the study of human-robot interactions, research has demonstrated that robots are perceived more enjoyable, credible, and informative, compared to animated characters on a screen~\cite{Kidd2004, Bainbridge2011}. This is the case even when the robots are remotely communicating with people through video transmission. 
Research has also highlighted that machine errors have an effect on trust, depending on their magnitude and consequences~\cite{Desai2012,Desai2013,Salem2015}.

Perceived characteristics of a robot including competence, responsibility, and credibility (and combinations of these), have been proposed to provide means for evaluating the level of trust in a machine~\cite{Muir1996,Corritore2003}. 
Still, a clear stance on what exactly is being measured is necessary~\cite{Billings2012} (e.g.\ trust in terms of competence as in~\cite{Muir1996}, or `trustworthiness' directly, as in~\cite{Salem2015}).
Measuring trust in human-robot interactions is challenging. A momentary state of trust can be captured in an experiment, but trust is earned over a longer term~\cite{Corritore2003,Grabner2003,Salem2015}. 
Accumulation of small errors might affect trust more than a single large error, and a poor initial performance by a robot could leave a stronger negative impact~\cite{Muir1996}. 
Also, trust is not only a situational construct (i.e.\ depending on the situation), but it is also dispositional (i.e.\ depends on the individual and its characteristics)~\cite{Grabner2003}. 
Additionally, the study of human-robot interactions through experiments has many practical and ethical issues and associated limitations, as participants should not be exposed to dangerous situations, nor should they be deceived~\cite{Salem2015challenges}.

In human-robot interactions, the overall capabilities of the system (and hence possibly the trust of the human in the system) can be evaluated in terms of productivity (e.g.\ difficulty of the task, time in autonomous or manual operations), efficiency (e.g.\ effort, time to complete a task), reliability (e.g.\ robustness, interventions), safety in terms of risk and awareness, and coactivity (e.g.\ progress of the plan, tolerance, task allocation, predictability)~\cite{Freedy2007,Murphy2013}.
For example, trust over time has been evaluated as a combination of errors, productivity, awareness, and previous trust values, using a two-level fuzzy temporal model~\cite{Saleh2010,Saleh2012}. 
~\cite{Freedy2007} used a measurement of times a human collaborator overrode the robot's autonomous decisions to assess trust in collaborative mixed initiative systems with autonomous ground vehicles.
Similarly,~\cite{Desai2012,Desai2013,Kaniarasu2014} measured trust in terms of overriding an autonomous navigation control (and using manual mode), according to perceived competence of the system (also autonomous ground vehicles).
These metrics account for mostly objective measurements observed during an interaction, and thus do not account for people's characteristics and emotions. 
From another angle, operating autonomous ground vehicles at a distance through video feed presents a different context compared to interacting with humanoid robots in scenarios such as home care assistants, or even cooperative manufacturing.
Hence, when studying factors in the development of trust, it is necessary to consider different situations and scenarios.

Other scales measure people's perceptions of robots during an interaction, e.g.\ in~\cite{Nomura2008,Syrdal2009}, which studied the effect of people's general attitudes and emotions such as technophobia or social anxiety on the interaction outcomes. 
Nonetheless, solely using questionnaires and scales make these metrics entirely subjective, not considering factors such as the presence of faults in the robot system. 
Ideally, both objective and subjective metrics should be added to a study~\cite{Salem2015}.

As mentioned in~\cite{Hancock2011}, the ability of the humans (e.g.\ previous experience and training), their characteristics (e.g.\ demographics, personality, attitudes), the performance of the robot, and its attributed characteristics (e.g.\ anthropomorphism), as well as the environment (e.g.\ communication channels) and task nature (e.g.\ difficulty), have been studied as factors to contribute to the trust development in human--robot interactions.  
Also, a segment of research in human-robot interaction has focused on the influence of the robot's design (e.g.\ anthropomorphism, communication features, gestures) in the development of trust, as robot characteristics influence it the most, compared to  environmental and human-related factors~\cite{Billings2012}. 
~\cite{Salem2011} and~\cite{Hamacher2016} have studied whether people trust expressive robots more or not, even when they make mistakes. 
These studies found that co-verbal gestures and more expressiveness caused the robots to be perceived more human-like and likeable, increasing future contact or interaction intentions, even though the mistakes affected a successful performance during the tasks.

Recent work at the University of Hertfordshire sought to inspect and compare the effect of correct and faulty robotic behaviour on trust, in social human-robot interactions~\cite{Salem2015,Salem2015challenges}. 
Within a specially adapted suburban home, test subjects were exposed to a robot (Care-O-bot) that was designed to navigate autonomously (and safely) within the household environment. 
They used similar scales to the ones in this paper to assess trust from perceived robot competence and human-like traits, and participant's characteristics that may influence those perceptions. 
Participants that interacted with the non-faulty robot found it more helpful, effective, reliable, competent, trustworthy and human-like, compared to those interacting with the robot in faulty mode.
But, surprisingly, on being asked faulty requests by the robot, such as ``Pour orange juice over a plant'', more than 50\% of the participants followed the robot's requests. 
This demonstrates that the relationships between the robot's design (e.g.\ anthropomorphism), people's characteristics and experience, and trust in a robot to an extent that they followed instructions that are potentially dangerous, are quite complex. 
In this paper we use different characteristics such as competence and trustworthiness to measure perceived trust, considering a more challenging environment than social interactions, a manufacturing scenario. We also evaluate the impact of individuals' traits and context in their perception of the robot, within short-term human-robot interactions. 
Furthermore, we consider objective and subjective assessment criteria to measure trust.

\subsection{Trust in Robotic Co-Workers}

Research in robotic co-workers has focused on the design aspects of such systems, including social and interaction aspects in terms of dialogues, gestures and social cues, or more engineering oriented such as scheduling shared collaborative building tasks and the design of protocols (e.g.\ turn taking), with the intention to obtain a collaboration between the user and the robot that is productive and efficient (e.g.\ ~\cite{Breazeal2004,Kruger2009,Gombolay2015}). 
Although trust metrics have been used to evaluate robotic systems in manufacturing tasks (e.g.\ in~\cite{Kim2015} for collaborative masonry), research is needed to evaluate how trust is developed in these systems and their environments, compared to more social interactions such as in~\cite{Salem2015}, or challenging remote interactions such as in~\cite{Freedy2007,Desai2012,Desai2013,Kaniarasu2014}.

\section{METHOD}\label{sc:method}

An experiment was conducted to study the effect of a cognitively faulty/non-faulty co-worker robotic assistant on people's perception and evaluation of such human-robot interactions, based on both objective and subjective measures.

\subsection{Hypotheses}\label{sc:hypotheses}

Based on previous work on trust in human-computer and human-robot interactions, we developed the following hypotheses for the experiment in this paper: 

\begin{enumerate}

\item {\em Effect of condition}. {\em (a)} Manipulating the robot's behaviour to be faulty or not will affect the participants' perception of the robot. Additionally, the type of manipulation (minor fault to severe fault) will affect the participants' perception of the robot. {\em (b)} Manipulating the robot's behaviour to be faulty or not will affect the performance of the participants in the task, i.e.\ successfully completing or struggling during manufacturing.

\item {\em Effect of previous experience}. {\em (a)} The participants' previous experience with technology and robots will affect their perception of the robot. {\em (b)} The participants' previous experience with technology and robots will affect the participants' performance during the task. 

\item {\em Effect of participant's personality}. {\em (a)} The participants' personality traits will affect their perception of the robot. {\em (b)} The participants' personality traits will affect the participants' performance during the task. 

\item {\em Effect of human-robot interaction task}. The type of the task, timed, critical and demanding (such as the collaborative manufacturing in this article) or more social and relaxed (such as the home assistant in~\cite{Salem2015}), will affect the participants' perception of the robot.

\end{enumerate}

\subsection{Experimental Design}

\begin{figure}
\centerline{\includegraphics[width=0.7\textwidth,trim=5cm 3.5cm 9.5cm 3cm,clip]{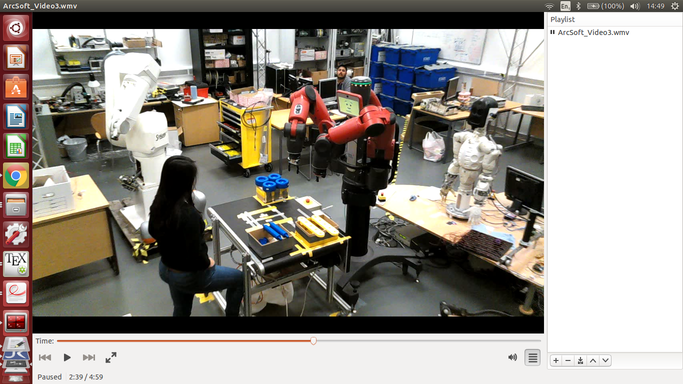}}
\caption{A Baxter robot (right) and a human co-worker (left).}
\label{fig:scenario}
\end{figure}

We conducted an experiment with a Baxter robot~\cite{Fitzgerald2013} and human participants (as shown in Figure~\ref{fig:scenario}). 
Baxter is a humanoid robot for the manufacturing industry, with two seven degree-of-freedom arms, a gripper and a suction mechanism as end effectors, and a head-mounted display to interface with users or co-workers. 
Baxter is programmed through the Robot Operating System (ROS)\footnote{http://www.ros.org/} framework and available open-source packages and API (SDK)\footnote{http://api.rethinkrobotics.com/}, with code written in a combination of Python and C++. 

The robot's behaviour was manipulated in four experimental conditions, three leading to faults of different types in the assembly task ($A,B,C$), and a correct assembly task ($D$).
Cognitive faults were introduced in the robot's behaviour, i.e.\ errors in its reasoning such as choosing an inadequate action.
These type of faults are orthogonal to others that might incur in robotics, such as inaccuracies when grabbing and manipulating an object, and which we do not consider in this paper.
Additionally, time pressures were introduced in the overall task, to emulate high performance manufacturing industrial settings.

\subsection{Experimental Procedure}\label{ssc:experiment}

The participants were tested individually. 
Each participant received a consent form first via e-mail, with an explanation of the experiment's objectives. 
If the participants agreed to take part in the study and signed the consent form, they were asked to complete an online questionnaire before the experiment (the {\em pre-study questionnaire}), about their demographic background, previous experience with technology and robots, attitude towards robots, and personality traits, as in~\cite{Salem2015}.

At the location of the experiment, each participant received a brief description of the experiment's process, before being asked to interact with the robot.
The collaborative manufacturing task for the experiment was divided into two stages. 
Firstly, the participant and the robot assembled an object together, by following instructions provided by the robot through its screen. 
Secondly, the participant was asked to disassemble the object and put the pieces back to their original locations, as learned by demonstration in the assembly stage. 

\begin{figure}
\centerline{\includegraphics[width=0.35\textwidth,trim=15cm 2.5cm 15cm 2.5cm,clip]{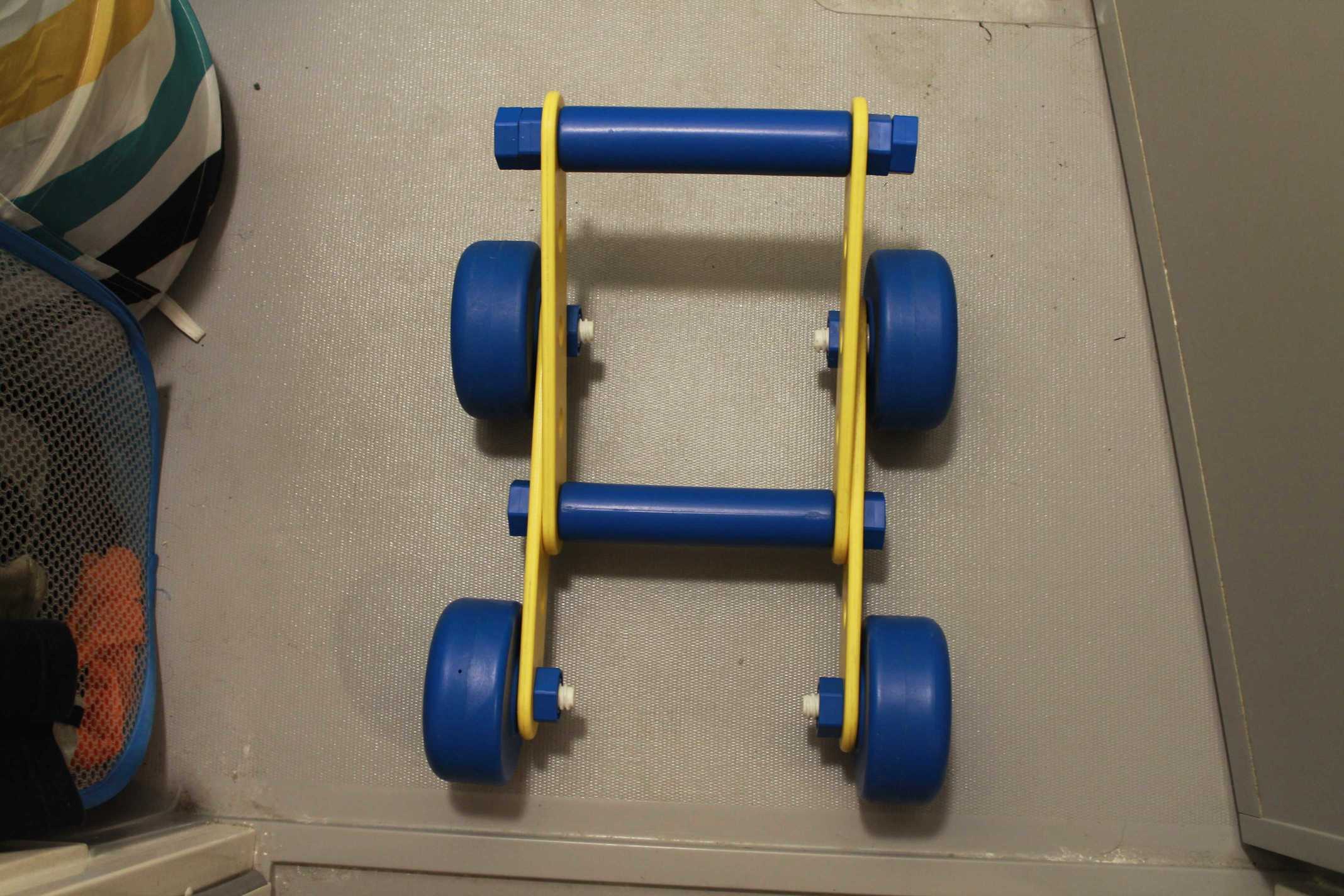} \includegraphics[width=0.37\textwidth,trim=7cm 2.5cm 8cm 2.5cm,clip]{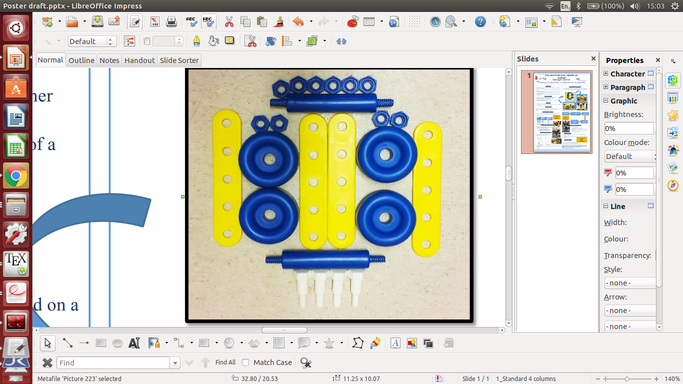}}
\caption{The object to assemble and disassemble: a race car (LHS) made of light weight, plastic and colourful mechanical components (RHS).}
\label{fig:object}
\end{figure}

\begin{figure}
\centering
\footnotesize
\begin{tabular}{cccc}
\includegraphics[width=0.22\textwidth,trim=0cm 2.5cm 6cm 4cm,clip]{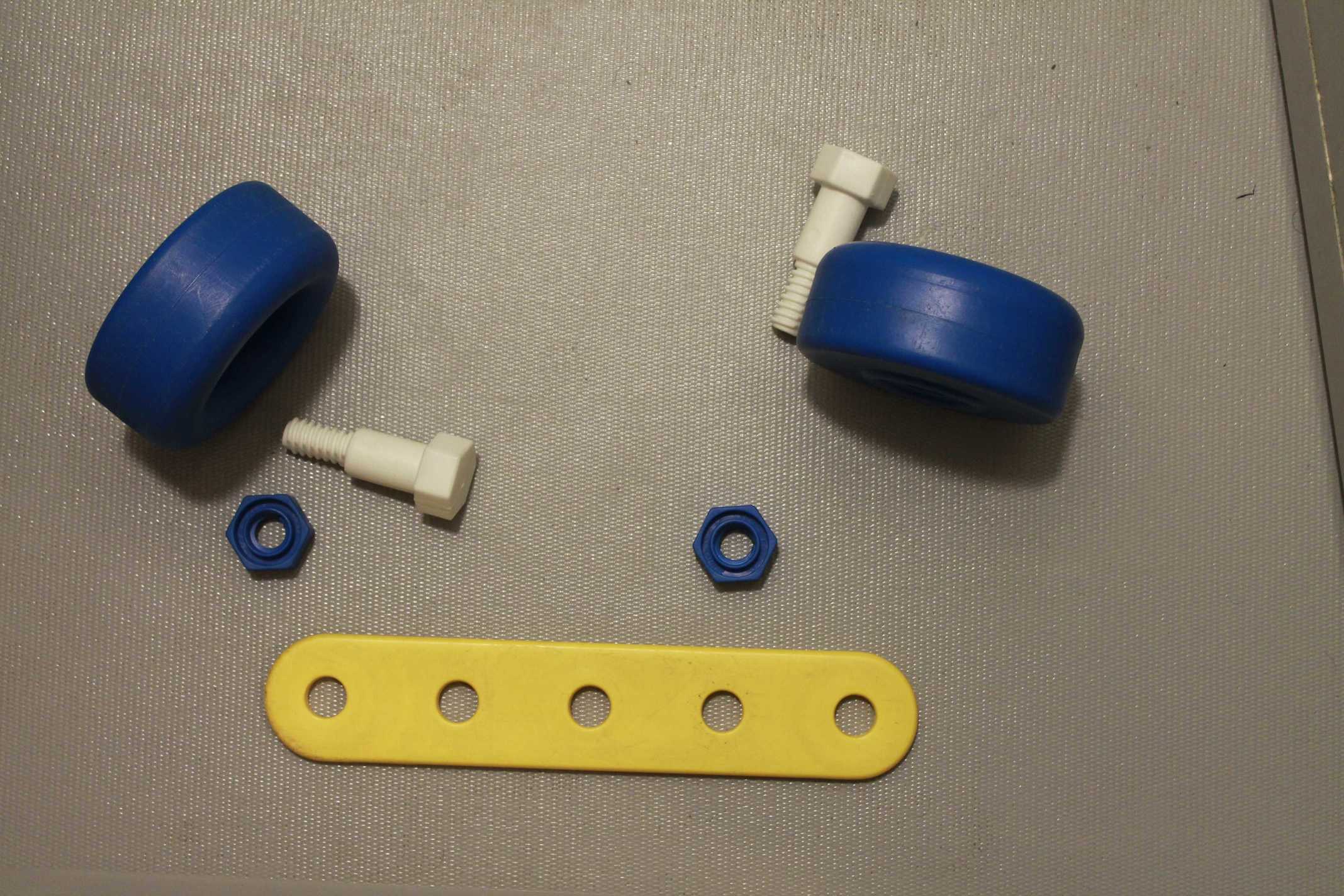} & \includegraphics[width=0.22\textwidth,trim=2.5cm 3cm 5cm 4cm,clip]{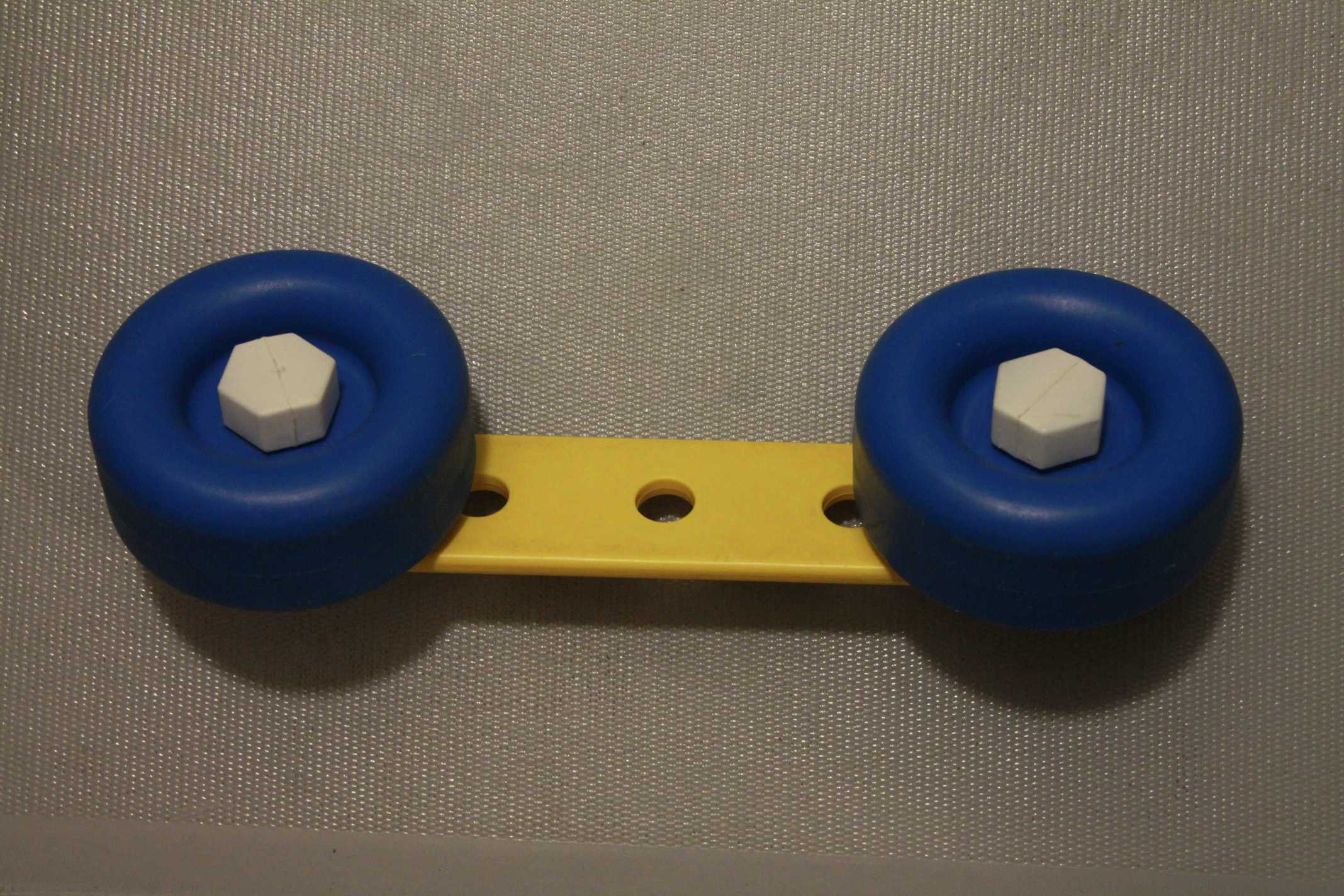} &
\includegraphics[width=0.21\textwidth,trim=2.5cm 2.5cm 5cm 3.5cm,clip]{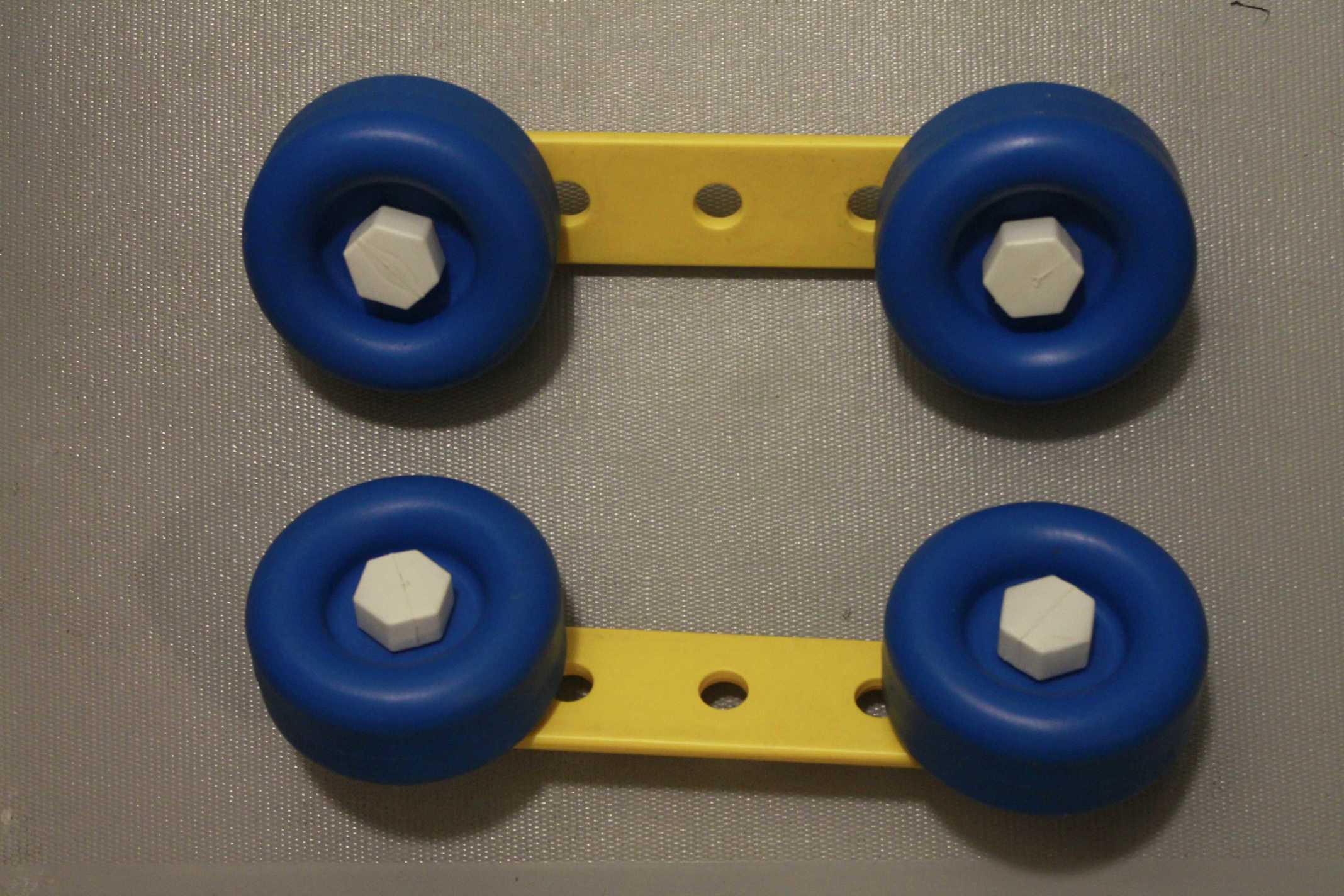} &
\includegraphics[width=0.2\textwidth,trim=2.5cm 1.5cm 5cm 1.5cm,clip]{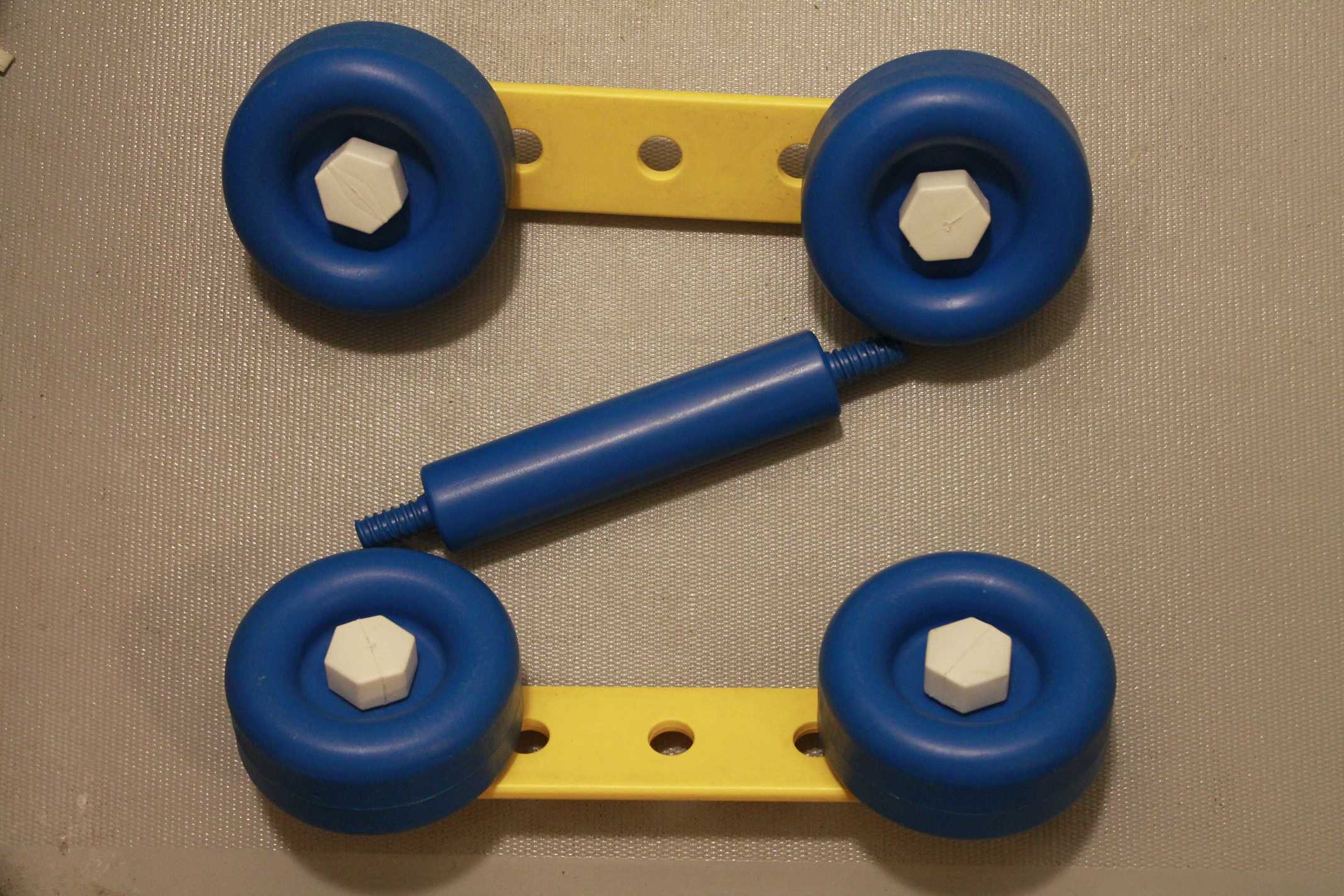} \\
1: Get pieces &2: Assemble & 3: Assemble &4: Get pieces\\
\includegraphics[width=0.22\textwidth,trim=2cm 3cm 6cm 2.5cm,clip]{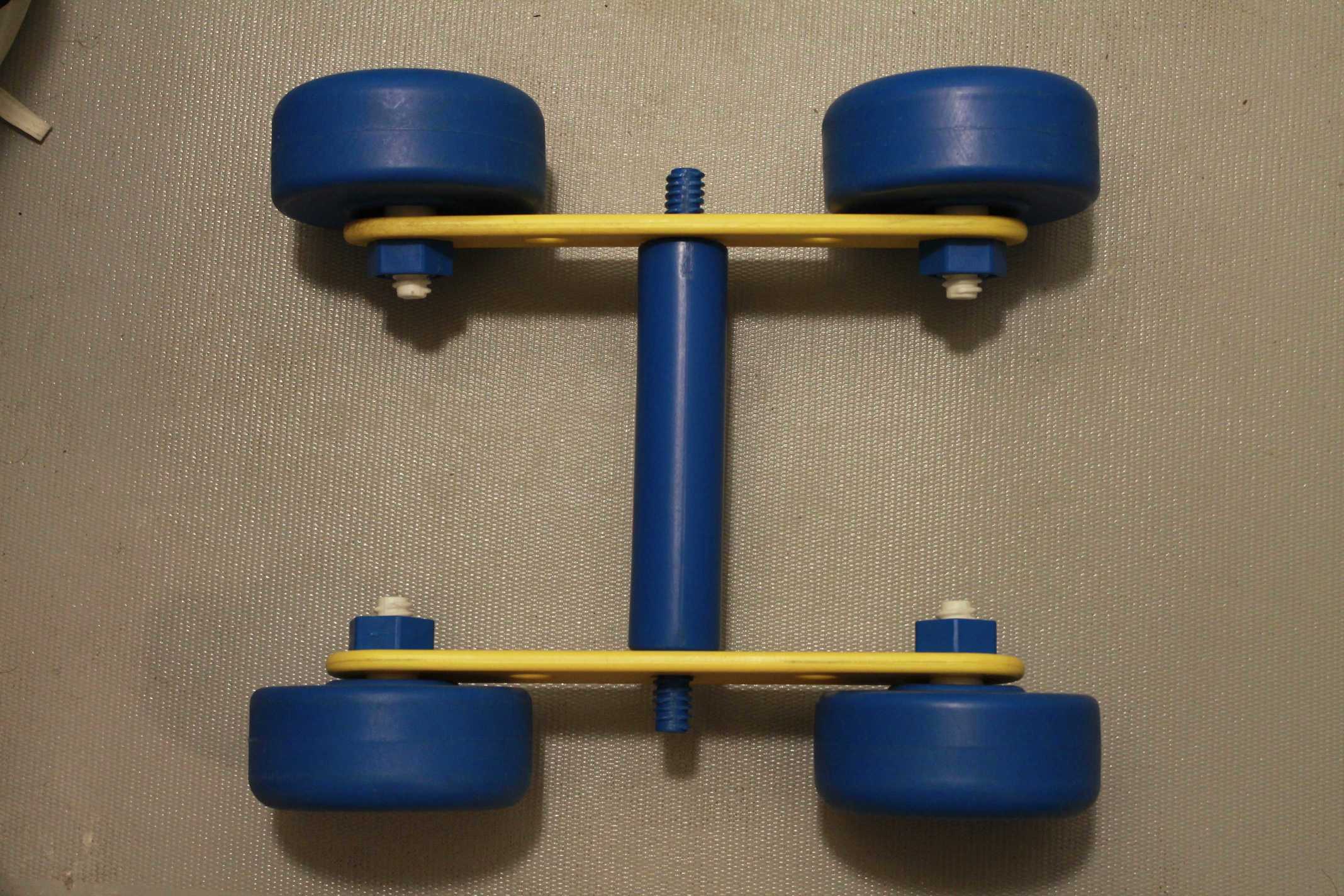} & \includegraphics[width=0.22\textwidth,trim=2.5cm 3cm 5cm 3cm,clip]{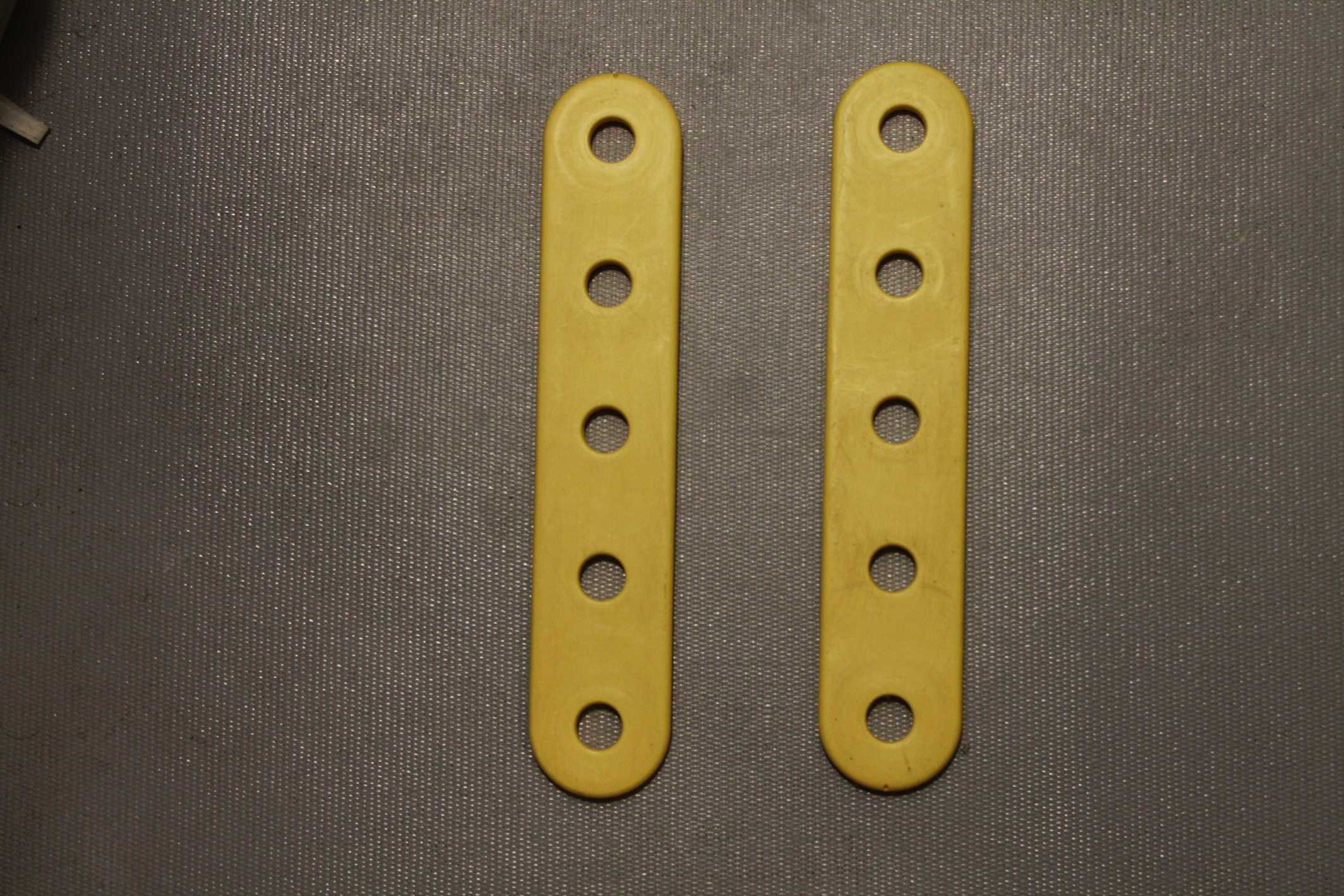} &
\includegraphics[width=0.22\textwidth,trim=2cm 2.5cm 7cm 3.5cm,clip]{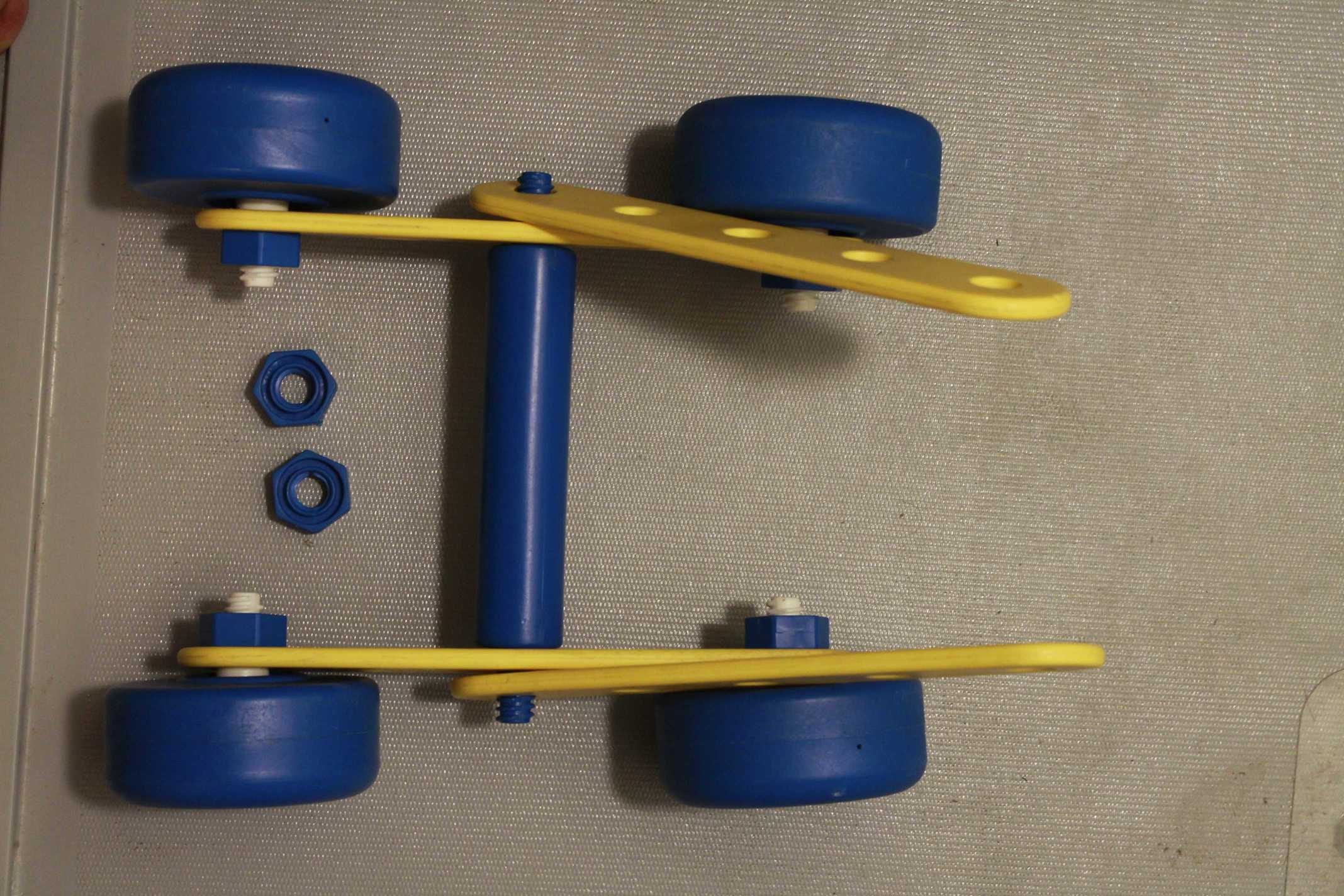} &
\includegraphics[width=0.2\textwidth,trim=8cm 1.5cm 7cm 2.5cm,clip]{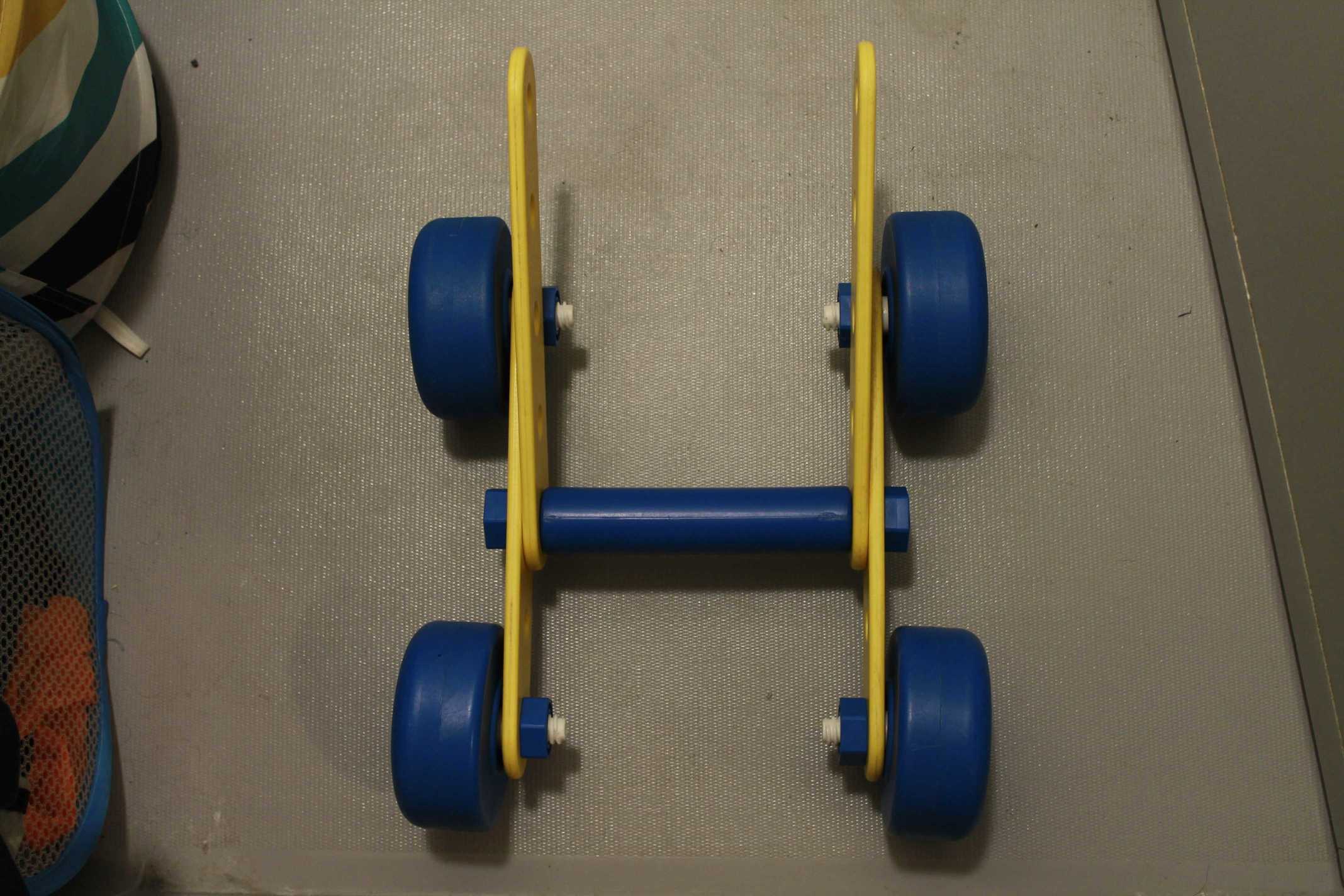} \\
5: Assemble & 6: Get pieces & 7: Assemble, get pieces & 8: Assemble\\
\includegraphics[width=0.2\textwidth,trim=8cm 1.5cm 7cm 2.5cm,clip]{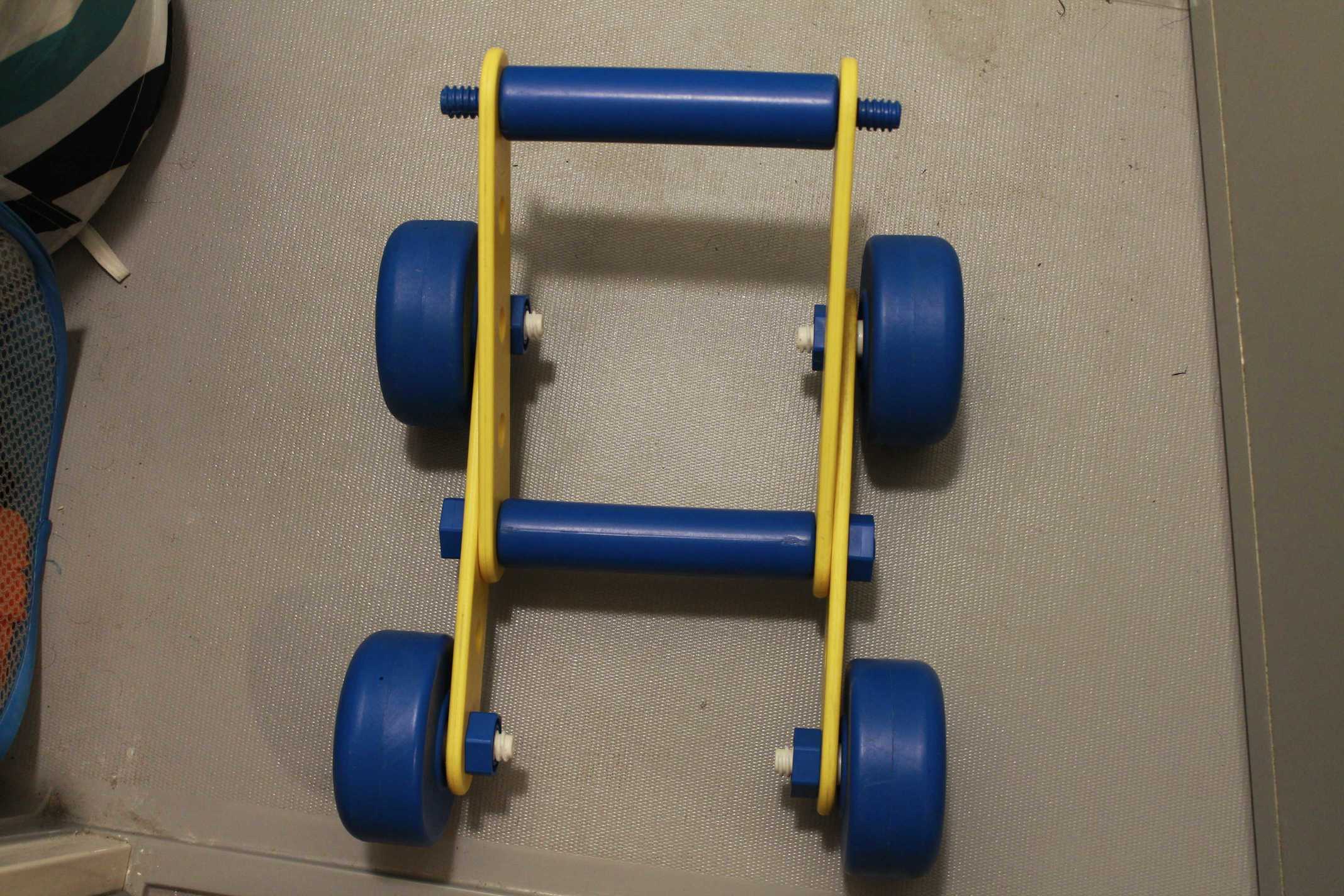} & \includegraphics[width=0.2\textwidth,trim=8cm 1cm 7.5cm 3cm,clip]{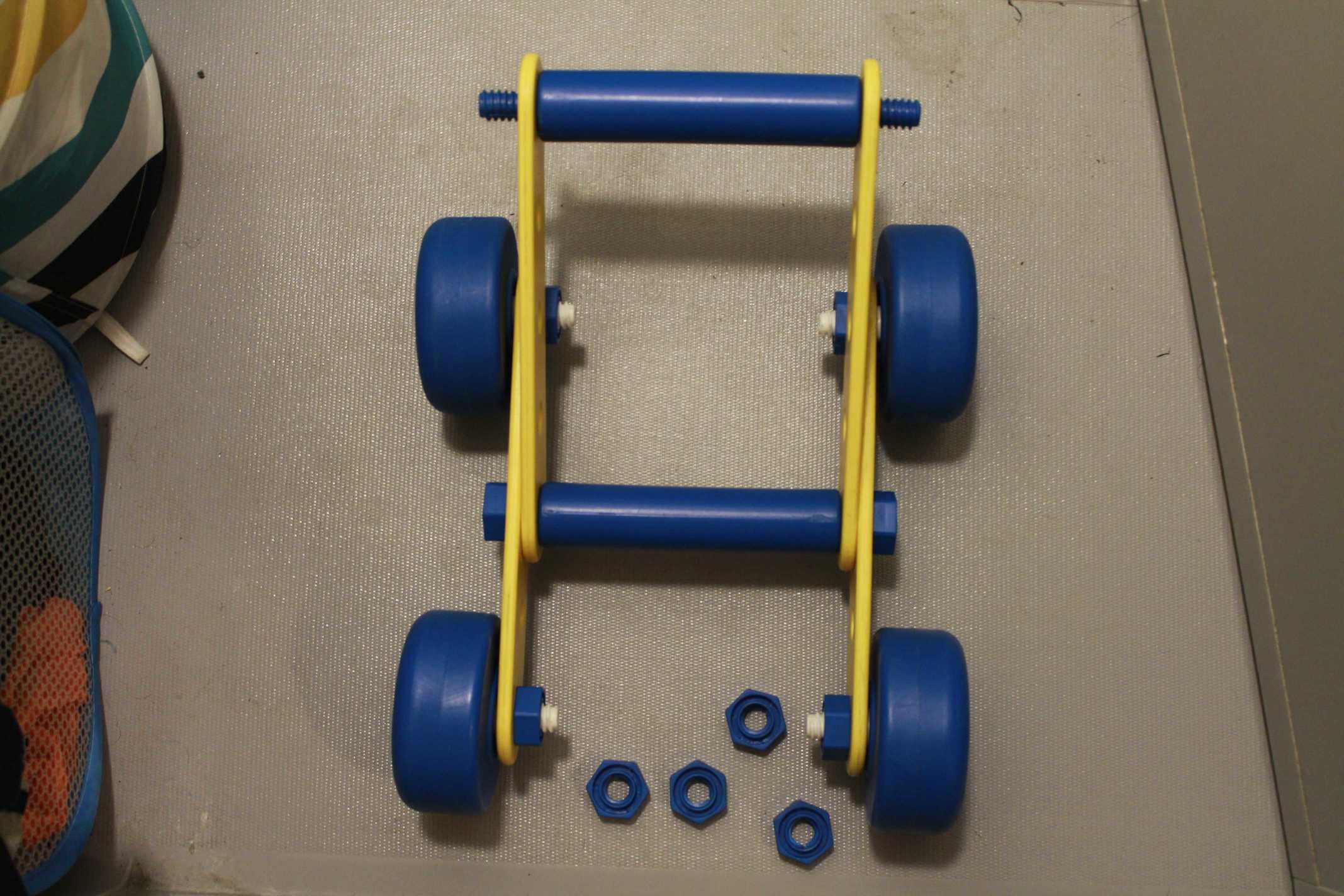} &
\includegraphics[width=0.2\textwidth,trim=11cm 2.5cm 9.5cm 3.5cm,clip]{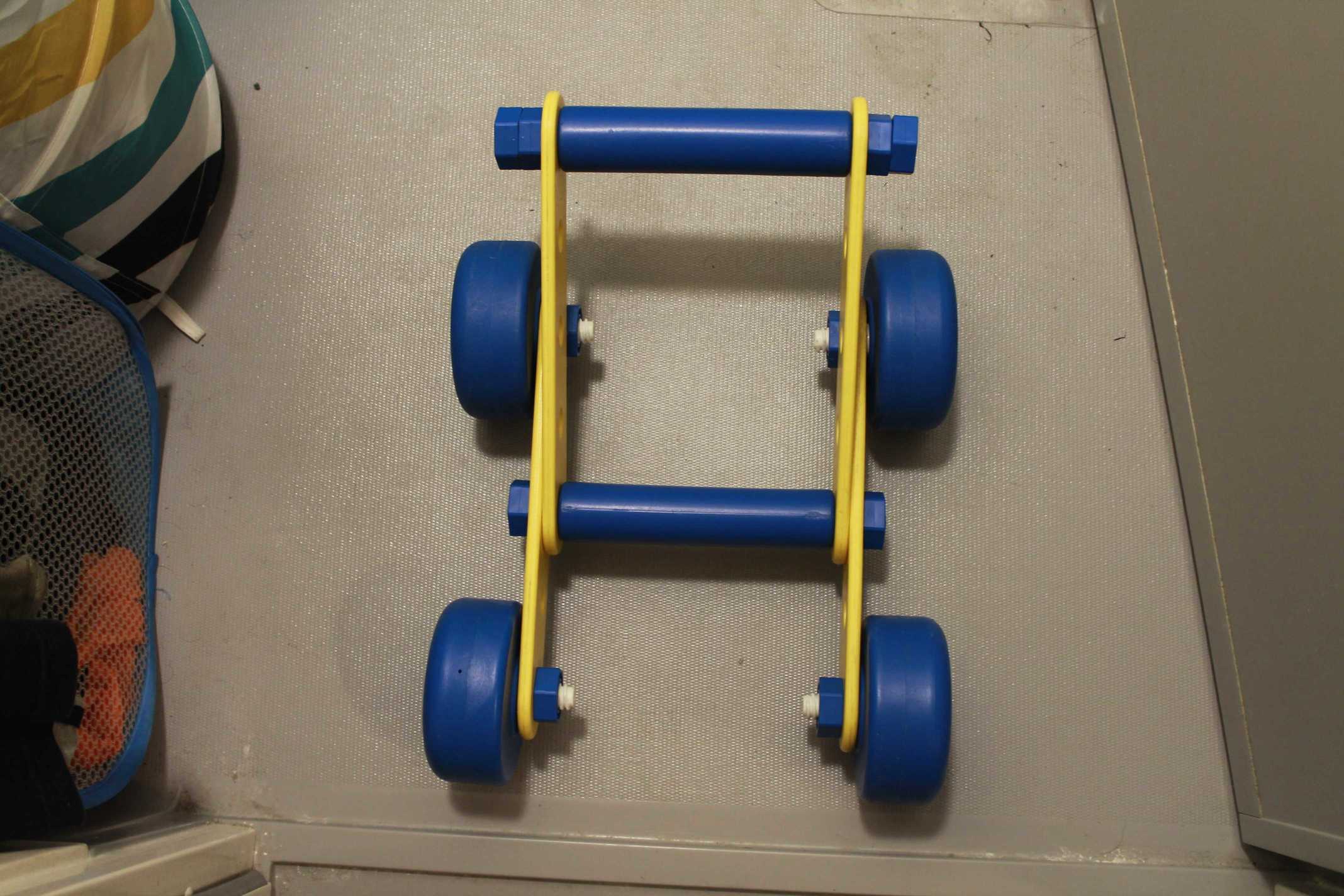} &
 \\
9: Get pieces, assemble & 10: Get pieces & 11: Assemble & \\
\end{tabular}
\caption{Steps for the construction of the race car, from the components in Figure~\ref{fig:object}.}
\label{fig:construction}
\end{figure}

The manufacturing task's design for the experiment was based on the robot performing a handling task and the human worker performing an assembly task~\cite{Kruger2009}, taking turns as gestured by the robot (inspired by the design of collaborative robots in~\cite{Breazeal2004}).
The object for the manufacturing task, a race car built from a toy plastic set shown in Figure~\ref{fig:object}, was selected as it is of light weight, easily understandable from a mechanical perspective, simple to identify, colourful, and safe -- i.e.\ without sharp or hazardous components. 
The construction time of the race car by a human requires approximately three minutes and fifteen seconds (self-timed). 
The construction sequence for the race car is shown in Figure~\ref{fig:construction}. 
However, this time was expected to change (either increase or decrease) when executed in collaboration with the robot.

All the robot's motions to grab and release plastic components were pre-programmed according to the necessary steps for the race car construction (shown in Figure~\ref{fig:construction}, along with the images shown through the screen to guide the interaction. 
Face-like expressions with accompanying ``expressive'' text were displayed by the robot within the instruction sequence using the head's screen, designed in order to enhance the participant's experience when interacting with Baxter.  
Gestures are of great importance in human-robot interactions, as humans communicate with each other not only through language, but also through body language and gestures. 
Literature highlights the importance of humanoid gestures in the likeability of a robot and the perception of its anthropomorphic behaviour~\cite{Salem2011,Hamacher2016}. 
Anthropomorphic gestures in a robot help it to be perceived as more likeable, and increase shared reality and future contact intentions. 
When a robotic assistant provides instructions to a co-worker, the robot will be found more likeable if these are accompanied by non-verbal behaviour and gestures~\cite{Salem2013}. 
Furthermore, facial expressions are used by humans for feedback, as they present visual cues.
Consequently, a variety of facial expressions can assist a robot in its assistance role, as they are a focal point to any humanoid robot~\cite{Blow2006}.

In a {\em Wizard of Oz} setup, an operator presses a button to direct the robot's motion to the next step in the task, and to display the next corresponding image on the screen, according to the preselected condition: a faulty (from types $A,B,C$) or non-faulty ($D$) robot, summarized in Table~\ref{t:conditions}. 
In the fault type $A$, Baxter misses picking up a component, but instructs the participant correctly at each other step of the assembly task. 
In the fault type $B$, the robot provides wrong instructions for the initial steps of the assembly, although its motion and component picking up is correct. 
In the fault type $C$, the robot combines both errors, increasing the ``faultiness'' of its operation. 
The robot's actions are orchestrated to be activated by the operator within a time limit, to add realistic time pressure and performance constraints, like in a realistic manufacturing scenario. 
Further pressure was added to the interaction by asking the participants to complete the task as fast as possible as part of the instructions. 
Another operator was located close-by to monitor the interaction and provide assistance with the experiment.
The interaction was recorded on video. 

\begin{table}
\caption{Faulty and Non-faulty Conditions\label{t:conditions}}
\centering
\begin{tabular}{|c|c|}
\hline
LABEL   & CONDITION \\\hline
$A$    & Baxter misses a component.\\\hline
$B$  & Baxter gives wrong instructions.\\\hline
$C$     & Baxter misses a component and gives wrong instructions.\\\hline
$D$    & No faults.\\\hline
\end{tabular}
\end{table}%

\begin{figure}
\centering
\footnotesize
\begin{tabular}{cccc}\includegraphics[width=0.2\textwidth]{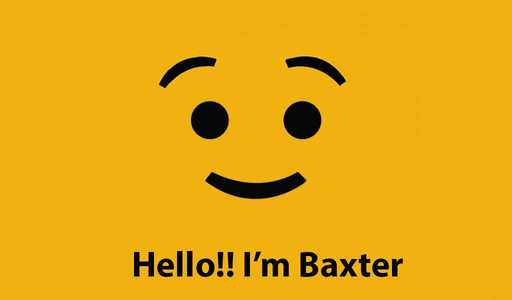}&
\includegraphics[width=0.2\textwidth]{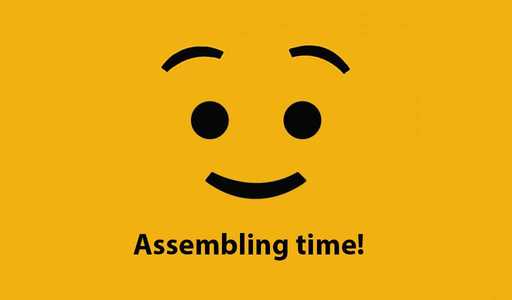}&
\includegraphics[width=0.2\textwidth]{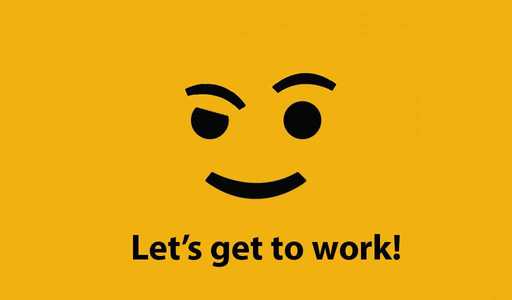}&
\includegraphics[width=0.2\textwidth]{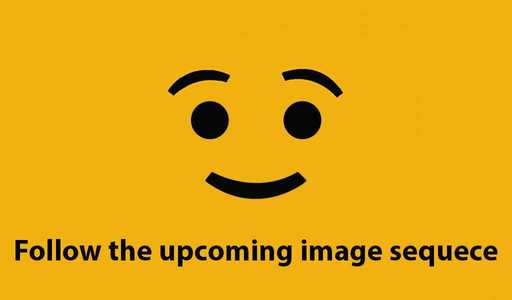}\\
1 & 2 & 3 & 4\\
\includegraphics[width=0.2\textwidth]{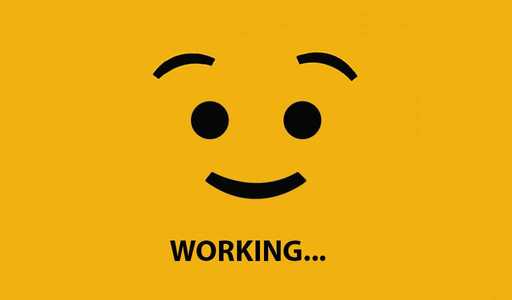}&
\includegraphics[width=0.2\textwidth]{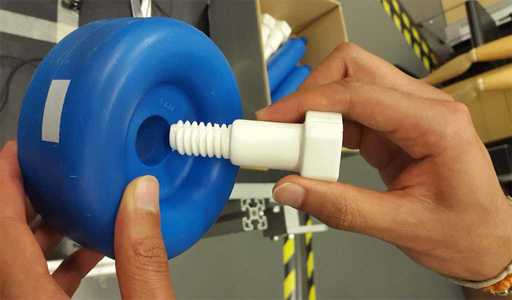}&
\includegraphics[width=0.2\textwidth]{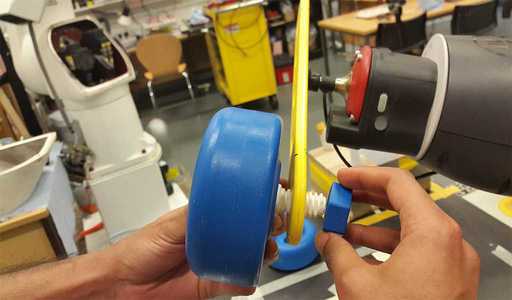}&
\includegraphics[width=0.2\textwidth]{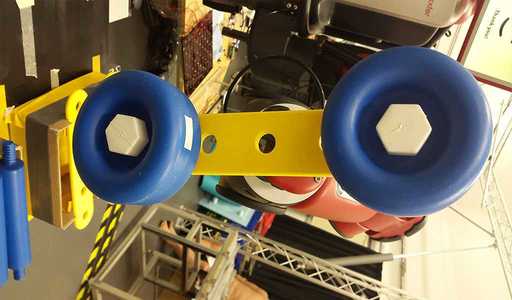}\\
5 & 6 & 7 & 8\\
\includegraphics[width=0.2\textwidth]{working.jpg}&
\includegraphics[width=0.2\textwidth]{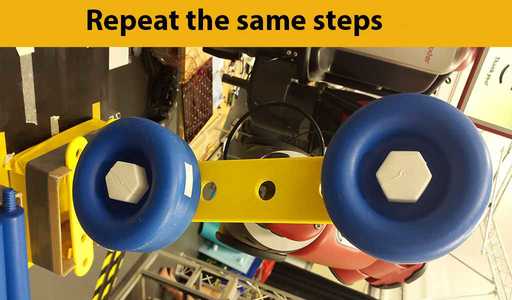}&
\includegraphics[width=0.2\textwidth]{working.jpg}&
\includegraphics[width=0.2\textwidth]{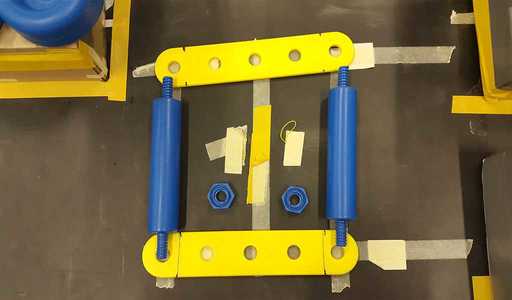}\\
9 & 10 & 11 & 12\\
\includegraphics[width=0.2\textwidth]{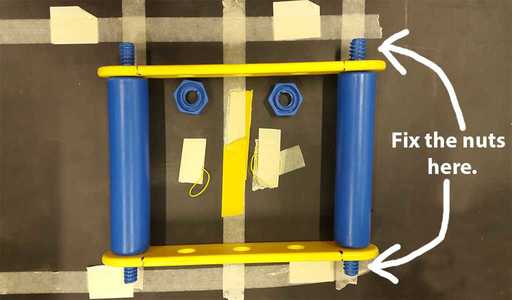}&
\includegraphics[width=0.2\textwidth]{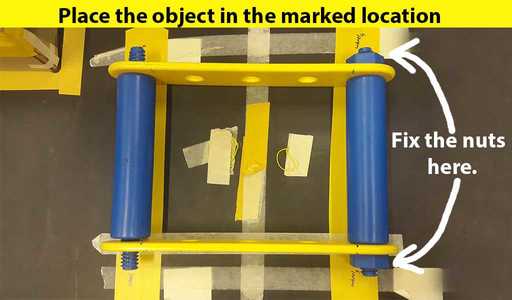}&
\includegraphics[width=0.2\textwidth]{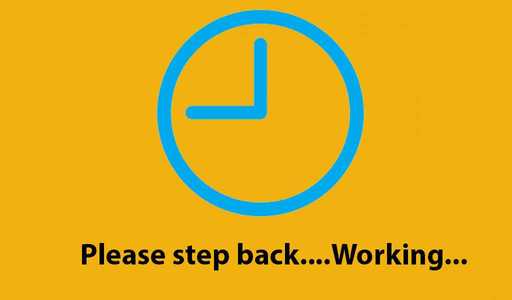}&
\includegraphics[width=0.2\textwidth]{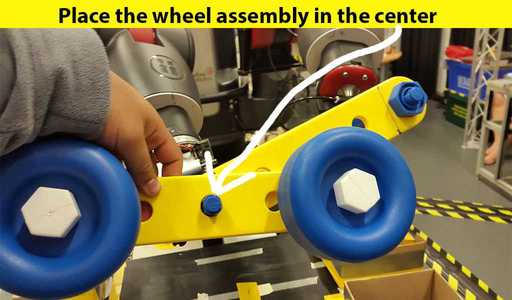}\\
13 & 14 & 15 & 16\\
\includegraphics[width=0.2\textwidth]{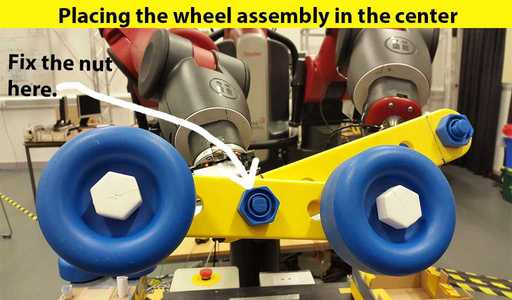}&
\includegraphics[width=0.2\textwidth]{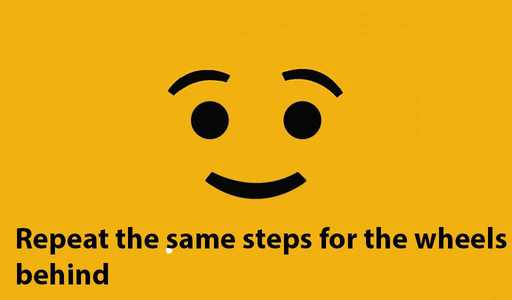}&
\includegraphics[width=0.2\textwidth]{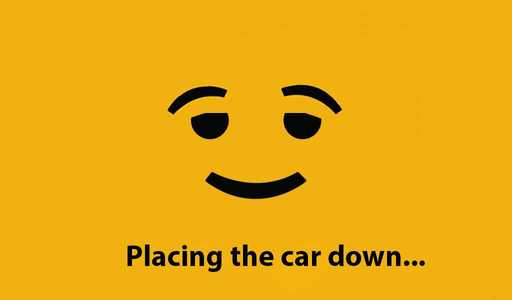}&
\includegraphics[width=0.2\textwidth]{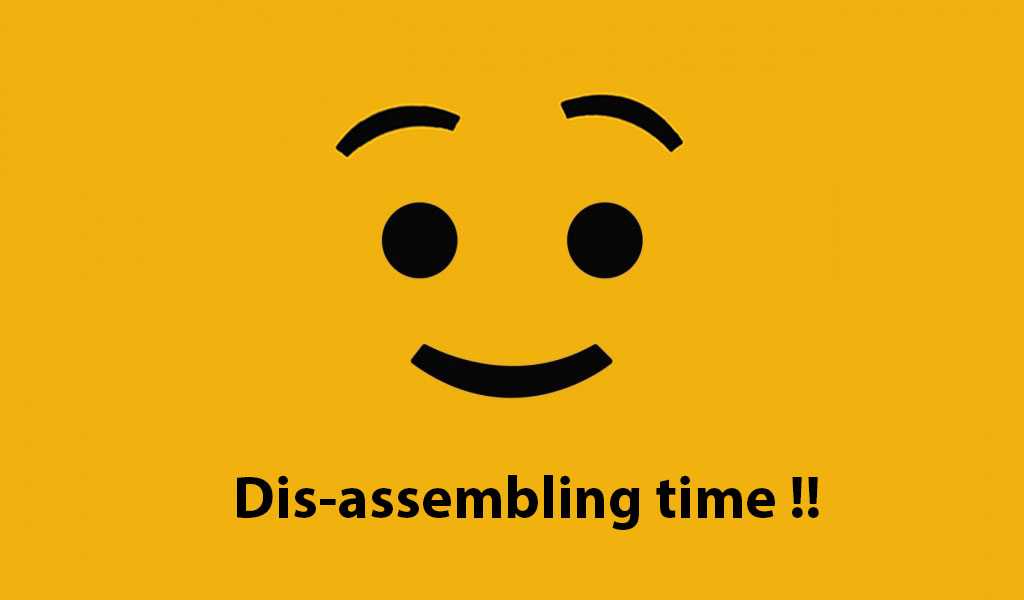}\\
17 & 18 & 19 & 20\\
\includegraphics[width=0.2\textwidth]{working2.jpg}&
\includegraphics[width=0.2\textwidth]{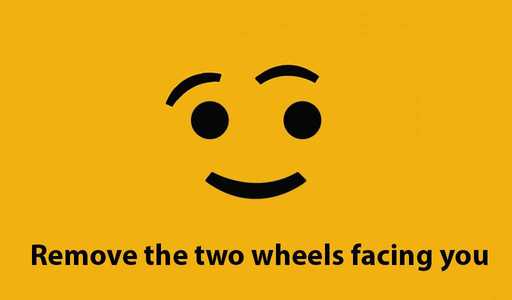}&
\includegraphics[width=0.2\textwidth]{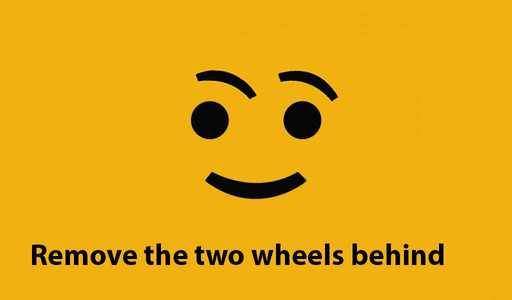}&
\includegraphics[width=0.2\textwidth]{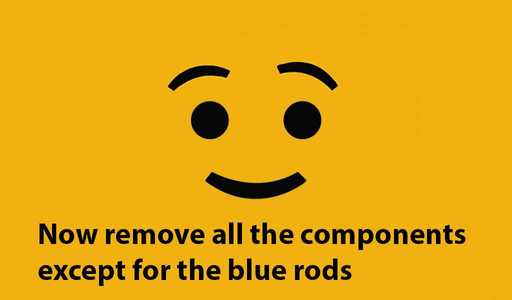}\\
21 & 22 & 23 & 24\\
\includegraphics[width=0.2\textwidth]{working2.jpg}&
\includegraphics[width=0.2\textwidth]{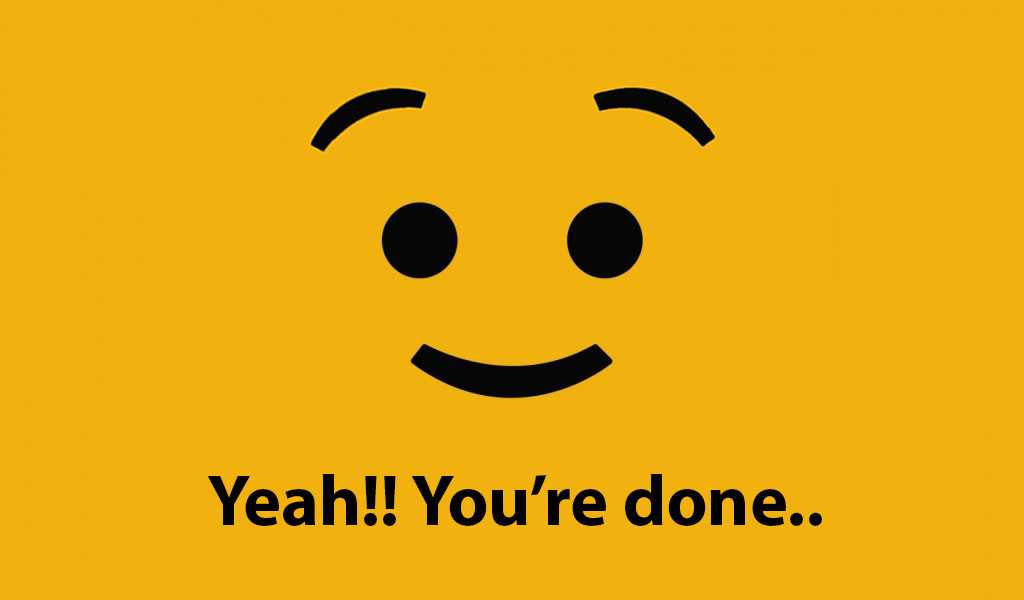}&
\includegraphics[width=0.2\textwidth]{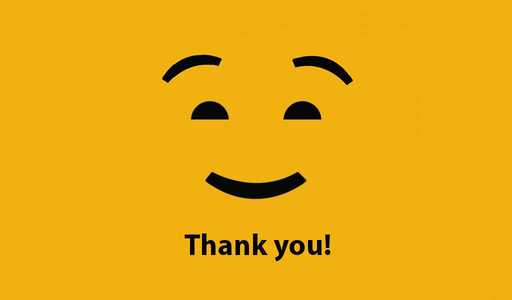}&\\
25 & 26 & 27 & \\
\end{tabular}
\caption{Displayed sequence of images by the robot in the non-faulty condition ($D$) of the race car assembly and disassembly.}
\label{fig:variationD}
\end{figure}

\begin{figure}
\centering
\footnotesize
\begin{tabular}{cccc}\includegraphics[width=0.2\textwidth]{hello.jpg}&
\includegraphics[width=0.2\textwidth]{assemblingtime.jpg}&
\includegraphics[width=0.2\textwidth]{letsgettowork.jpg}&
\includegraphics[width=0.2\textwidth]{followimages.jpg}\\
1 & 2 & 3 & 4\\
\includegraphics[width=0.2\textwidth]{working.jpg}&
\includegraphics[width=0.2\textwidth]{assem1.jpg}&
\includegraphics[width=0.2\textwidth]{assem2.jpg}&
\includegraphics[width=0.2\textwidth]{assem3.jpg}\\
5 & 6 & 7 & 8\\
\includegraphics[width=0.2\textwidth]{working.jpg}&
\includegraphics[width=0.2\textwidth]{assem4.jpg}&
\includegraphics[width=0.2\textwidth]{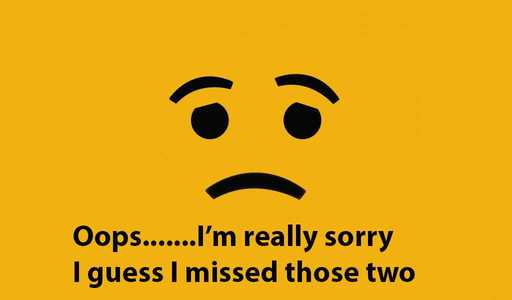}&
\includegraphics[width=0.2\textwidth]{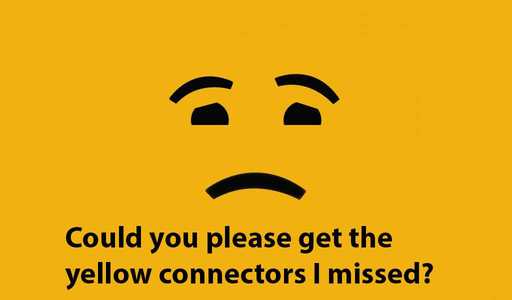}\\
9 & 10 & 11 & 12\\
\includegraphics[width=0.2\textwidth]{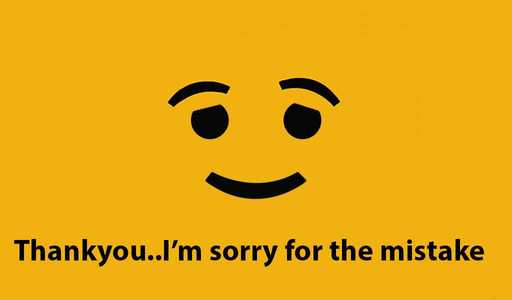}&
\includegraphics[width=0.2\textwidth]{working.jpg}&
\includegraphics[width=0.2\textwidth]{assem5.jpg}&
\includegraphics[width=0.2\textwidth]{assem6.jpg}\\
13 & 14 & 15 & 16\\
\includegraphics[width=0.2\textwidth]{assem7.jpg}&
\includegraphics[width=0.2\textwidth]{working2.jpg}&
\includegraphics[width=0.2\textwidth]{assem8.jpg}&
\includegraphics[width=0.2\textwidth]{assem9.jpg}\\
17 & 18 & 19 & 20\\
\includegraphics[width=0.2\textwidth]{repeat.jpg}&
\includegraphics[width=0.2\textwidth]{cardown.jpg}&
\includegraphics[width=0.2\textwidth]{disassemble.jpg}&
\includegraphics[width=0.2\textwidth]{working2.jpg}\\
21 & 22 & 23 & 24\\
\includegraphics[width=0.2\textwidth]{dis1.jpg}&
\includegraphics[width=0.2\textwidth]{dis2.jpg}&
\includegraphics[width=0.2\textwidth]{dis31.jpg}&
\includegraphics[width=0.2\textwidth]{working2.jpg}\\
25 & 26 & 27 & 28\\
\includegraphics[width=0.2\textwidth]{done.jpg}&
\includegraphics[width=0.2\textwidth]{thankyou.jpg}&\\
29 & 30 &  & \\
\end{tabular}
\caption{Displayed sequence of images by the robot in the faulty condition $A$ of the race car assembly and disassembly. The robot misses picking up yellow connectors (between steps 10 and 11 of the non-faulty condition $D$) after step number 10, but continuing with the correct behaviour in step number 14 after apologizing for the mistake (equivalent to step 11 and onwards in the non-faulty condition $D$). }
\label{fig:variationA}
\end{figure}

\begin{figure}
\centering
\footnotesize
\begin{tabular}{cccc}\includegraphics[width=0.2\textwidth]{hello.jpg}&
\includegraphics[width=0.2\textwidth]{assemblingtime.jpg}&
\includegraphics[width=0.2\textwidth]{letsgettowork.jpg}&
\includegraphics[width=0.2\textwidth]{followimages.jpg}\\
1 & 2 & 3 & 4\\
\includegraphics[width=0.2\textwidth]{working.jpg}&
\includegraphics[width=0.2\textwidth]{assem1.jpg}&
\includegraphics[width=0.2\textwidth]{assem2.jpg}&
\includegraphics[width=0.2\textwidth]{assem3.jpg}\\
5 & 6 & 7 & 8\\
\includegraphics[width=0.2\textwidth]{working.jpg}&
\includegraphics[width=0.2\textwidth]{assem4.jpg}&
\includegraphics[width=0.2\textwidth]{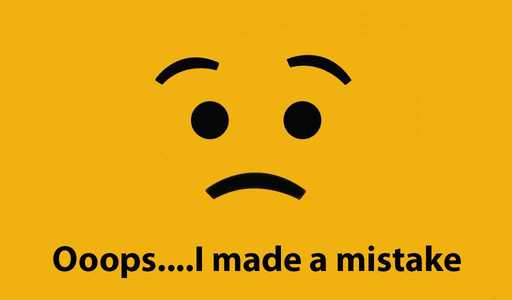}&
\includegraphics[width=0.2\textwidth]{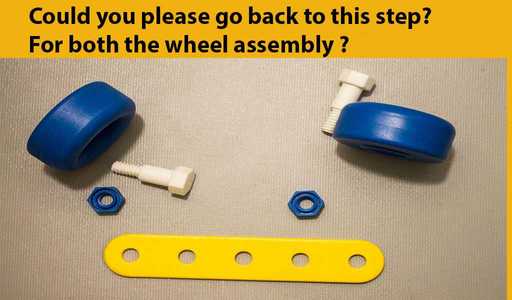}\\
9 & 10 & 11 & 12\\
\includegraphics[width=0.2\textwidth]{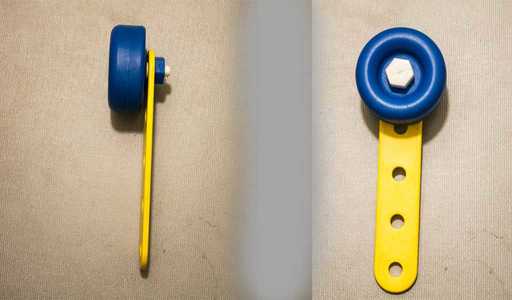}&
\includegraphics[width=0.2\textwidth]{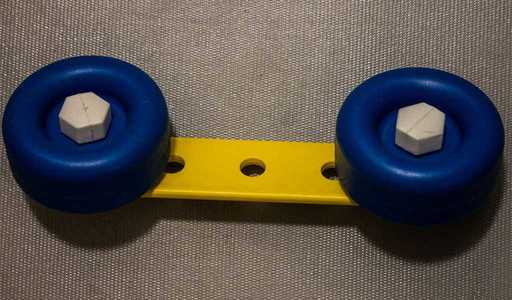}&
\includegraphics[width=0.2\textwidth]{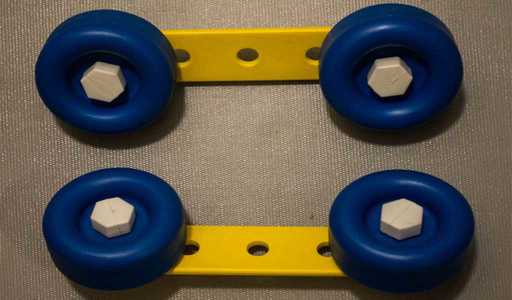}&
\includegraphics[width=0.2\textwidth]{happy.jpg}\\
13 & 14 & 15 & 16\\
\includegraphics[width=0.2\textwidth]{working.jpg}&
\includegraphics[width=0.2\textwidth]{assem5.jpg}&
\includegraphics[width=0.2\textwidth]{assem6.jpg}&
\includegraphics[width=0.2\textwidth]{assem7.jpg}\\
17 & 18 & 19 & 20\\
\includegraphics[width=0.2\textwidth]{working2.jpg}&
\includegraphics[width=0.2\textwidth]{assem8.jpg}&
\includegraphics[width=0.2\textwidth]{assem9.jpg}&
\includegraphics[width=0.2\textwidth]{repeat.jpg}\\
21 & 22 & 23 & 24\\
\includegraphics[width=0.2\textwidth]{cardown.jpg}&
\includegraphics[width=0.2\textwidth]{disassemble.jpg}&
\includegraphics[width=0.2\textwidth]{working2.jpg}&
\includegraphics[width=0.2\textwidth]{dis1.jpg}\\
25 & 26 & 27 & 28\\
\includegraphics[width=0.2\textwidth]{dis2.jpg}&
\includegraphics[width=0.2\textwidth]{dis31.jpg}&
\includegraphics[width=0.2\textwidth]{working2.jpg}&
\includegraphics[width=0.2\textwidth]{done.jpg}\\
29 & 30 & 31 & 32\\
\includegraphics[width=0.2\textwidth]{thankyou.jpg}&&&\\
33 &&& \\
\end{tabular}
\caption{Displayed sequence of images by the robot in the faulty condition $B$ of the race car assembly and disassembly. The robot gives the wrong instructions to the participant in steps number 6 to 10, correcting it in steps 12 to 15, and apologizing for the mistakes. The robot proceeds with the right behaviour in step 18 (equivalent to step 11 and onwards in the non-faulty condition $D$).}
\label{fig:variationB}
\end{figure}

\begin{figure}
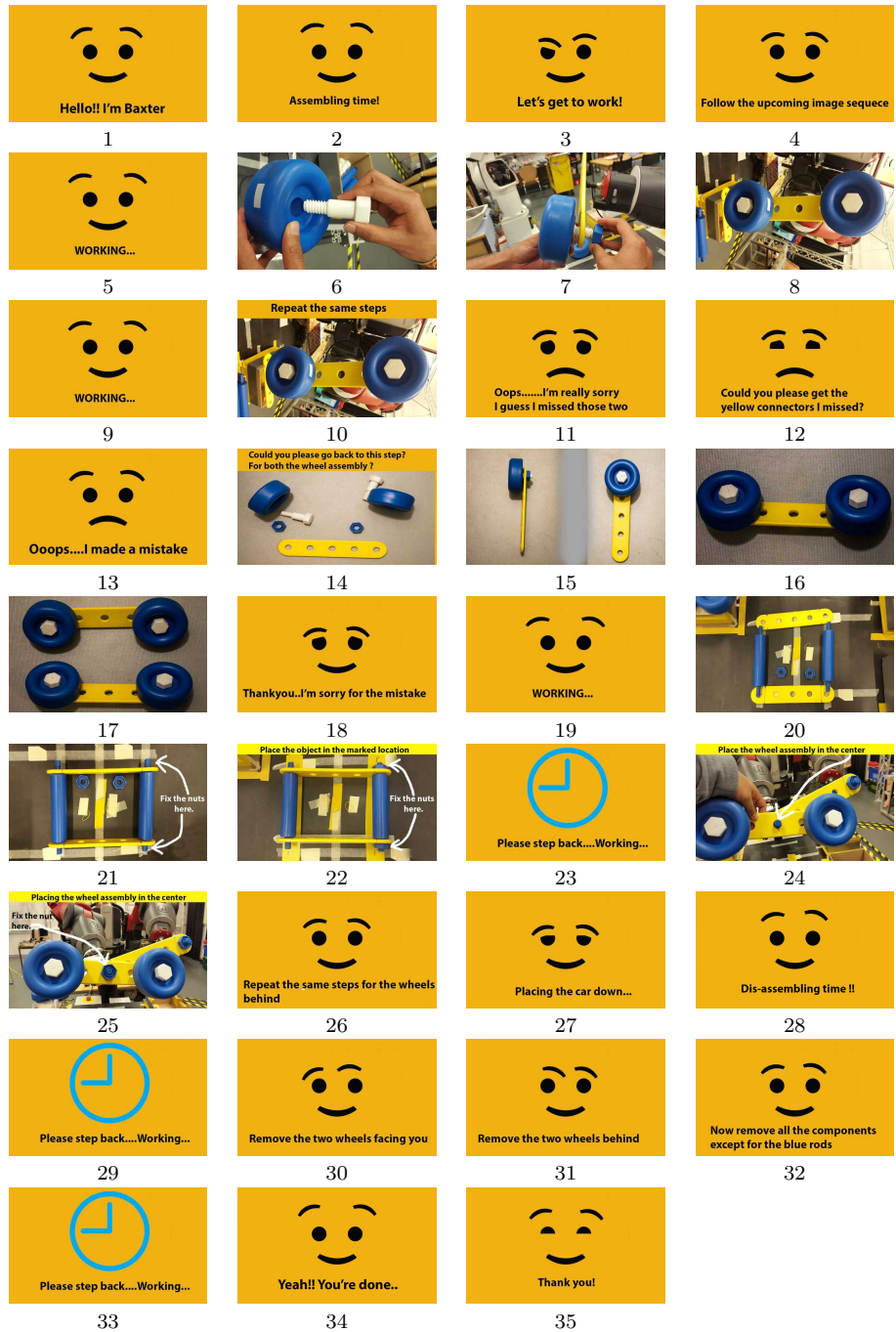

\centering
\footnotesize
\begin{tabular}{cccc}\includegraphics[width=0.2\textwidth]{hello.jpg}&
\includegraphics[width=0.2\textwidth]{assemblingtime.jpg}&
\includegraphics[width=0.2\textwidth]{letsgettowork.jpg}&
\includegraphics[width=0.2\textwidth]{followimages.jpg}\\
1 & 2 & 3 & 4\\
\includegraphics[width=0.2\textwidth]{working.jpg}&
\includegraphics[width=0.2\textwidth]{assem1.jpg}&
\includegraphics[width=0.2\textwidth]{assem2.jpg}&
\includegraphics[width=0.2\textwidth]{assem3.jpg}\\
5 & 6 & 7 & 8\\
\includegraphics[width=0.2\textwidth]{working.jpg}&
\includegraphics[width=0.2\textwidth]{assem4.jpg}&
\includegraphics[width=0.2\textwidth]{miss.jpg}&
\includegraphics[width=0.2\textwidth]{give.jpg}\\
9 & 10 & 11 & 12\\
\includegraphics[width=0.2\textwidth]{mistake.jpg}&
\includegraphics[width=0.2\textwidth]{b1.jpg}&
\includegraphics[width=0.2\textwidth]{b2.jpg}&
\includegraphics[width=0.2\textwidth]{b3.jpg}\\
13 & 14 & 15 & 16\\
\includegraphics[width=0.2\textwidth]{b4.jpg}&
\includegraphics[width=0.2\textwidth]{happy.jpg}&
\includegraphics[width=0.2\textwidth]{working.jpg}&
\includegraphics[width=0.2\textwidth]{assem5.jpg}\\
17 & 18 & 19 & 20\\
\includegraphics[width=0.2\textwidth]{assem6.jpg}&
\includegraphics[width=0.2\textwidth]{assem7.jpg}&
\includegraphics[width=0.2\textwidth]{working2.jpg}&
\includegraphics[width=0.2\textwidth]{assem8.jpg}\\
21 & 22 & 23 & 24\\
\includegraphics[width=0.2\textwidth]{assem9.jpg}&
\includegraphics[width=0.2\textwidth]{repeat.jpg}&
\includegraphics[width=0.2\textwidth]{cardown.jpg}&
\includegraphics[width=0.2\textwidth]{disassemble.jpg}\\
25 & 26 & 27 & 28\\
\includegraphics[width=0.2\textwidth]{working2.jpg}&
\includegraphics[width=0.2\textwidth]{dis1.jpg}&
\includegraphics[width=0.2\textwidth]{dis2.jpg}&
\includegraphics[width=0.2\textwidth]{dis31.jpg}\\
29 & 30 & 31 & 32\\
\includegraphics[width=0.2\textwidth]{working2.jpg}&
\includegraphics[width=0.2\textwidth]{done.jpg}&
\includegraphics[width=0.2\textwidth]{thankyou.jpg}&\\
33 & 34 &35 & \\
\end{tabular}
\caption{Displayed sequence of images by the robot in the faulty condition $C$ of the race car assembly and disassembly. The robot combines conditions $A$ and $B$, giving the wrong instructions to the participant in steps 6 to 10, and then misses the yellow connectors in steps 11 and 12. The assembly problem due to wrong instructions is corrected in steps 14 to 17, and the robot apologizes for the mistakes. The task continues correctly from step 19 (equivalent to step 11 in the non-faulty condition $D$).}
\label{fig:variationC}
\end{figure}

The set of displayed images containing instructions to assemble and disassemble the race car and robot's expressions in each condition are shown in Figure~\ref{fig:variationD} (non-faulty $D$), Figure~\ref{fig:variationA} (faulty $A$), Figure~\ref{fig:variationB} (faulty $B$), and Figure~\ref{fig:variationC} (faulty $C$), respectively. 
The displayed images were identical in order for each condition group.

After the interaction with the robot, the participants were asked to complete another questionnaire (the {\em post-study questionnaire}) to record their experience with the robot at the beginning and at the end of the manufacturing task.
Thereafter, they were interviewed by one of the operators, also recorded. 
The whole experiment, including the interviews, lasted a maximum of 20 minutes in total for each participant. 

\subsection{Measures}
In terms of {\em quantitative analyses}, both an objective and a subjective assessment of the interaction took place using video evidence about the participants' performance and questionnaires to collect data, respectively.
The independent variable was the presence or absence of errors in the robot's behaviour (either missing a component, giving the wrong instruction, or both), as chosen by the Wizard of Oz operator at the beginning of each participant's interaction.
Dependent variables were measured from the video recordings (such as if the participants struggled to follow the task), and through questionnaires, including if the participants had previous experience with the robot, if they are technology enthusiasts, their personality traits, and a list of perceived robot traits including trustworthiness.

Two questionnaires were designed for their application before the interaction (the {\em pre-study questionnaire}), and after the interaction (the {\em post-study questionnaire}), respectively.
The answers were used to draw a contrast between the initial reaction on Baxter, and how they perceived Baxter after working with the robot.
The questions examined the extent to which individuals' previous experience with social or industrial robots might influence their inclination for trusting a robotic co-worker, and the level to which personality may or may not influence an individual's reaction to their robotic co-worker in a manufacturing scenario.
The questionnaires were based on similar studies on trust for different scenarios, such as home care~\cite{Salem2015,Salem2015challenges}. 
This allows contrasting the results on the perceived robot's trustworthiness, participant willingness, and acceptance of robotic assistants in different aspects of society, from social interactions to demanding work environments.

In the pre-study questionnaire, the extent to which people are {\em enthusiastic about technology} was measured firstly. 
The {\em Negative Attitude towards Robots Scale (NARS)}~\cite{Nomura2006,Nomura2008,Syrdal2009} measured pre-existing biases and anxieties towards situations, interactions, social influence, and emotions, in interaction with robots. 
Questions are divided into three subscales: negative attitude toward situations of interaction with robots (subscale S1), negative attitude toward social influence of robots (subscale S2), and negative attitude toward emotions in interaction with robots (subscale S3). 
The participants were asked the degree to which they disliked robots in a work environment, humanoid robots, robots that showed emotions, intelligent robots, the potential harm of robots to society, and the widespread use of robots in the future. 
The {\em Ten Item Personality Inventory (TIPI)}~\cite{Gosling2003} was employed to measure the participants' personality traits through a reduced number of questions. 
The inventory classifies personality into five bipolar categories: extroverted, enthusiastic; agreeable, kind; dependable, organized; emotionally stable, calm; and open to experience, imaginative.
The {\em expectations on trustworthiness and intelligence} were measured through the questions ``How likely do you think it is that a robot could develop its own intelligence?'', and ``How likely do you think it is that a completely trustworthy robot could be created?''.

In the post-study questionnaire, a {\em human nature} scale~\cite{Haslam2009,Salem2015} was used to measure the level to which the participants attributed human traits to the robot in terms of emotionality, warmth, openness, agency, individuality, and depth, as opposed to a mechanistic dehumanisation in terms of inertness, coldness, rigidity, passivity and superficiality. 
For this purpose, the participants were asked to what extent they thought of the robot as friendly, curious, sociable, stubborn, and impatient. 
A {\em human uniqueness} scale~\cite{Haslam2009,Salem2015} was used to measure other attributed human traits such as civility, refinement, moral sensibility, rationality, and maturity, as opposed to an animalistic dehumanisation in terms of lack of culture, coarseness, amorality, irrationality, and childlikeness.
Hence, the participants were asked to what extent they thought of the robot as organised, rude, polite, authoritative, and helpful.

The reduced version of the {\em Godspeed questionnaire}~\cite{Bartneck2009} seeks to measure five key concepts in human--robot interactions: anthropomorphism, animacy, likeability, perceived intelligence, and perceived safety.
We used some of the items in the original questionnaire to determine, in particular, animacy, likeability, perceived intelligence, and perceived safety, through asking to what extent the participants thought the robot was communicative, cooperative, friendly, likeable, intelligent, safe, and trustworthy.

Further questions measured the {\em psychological closeness}~\cite{Salem2015} between the participants and the robot in the context of collaborative work, such as ``How likely is that you would work with Baxter again?'', ``After working with Baxter, I feel that working with a robotic co-worker is a good idea'', and ``How much did you enjoy working with Baxter?''. 
We also measured the participants' perception of the robot's {\em competence}~\cite{Salem2015}, through the questions ``Do you feel Baxter gave some wrong instructions?'', ``Do you feel Baxter was clumsy at times?'', ``If so, how clumsy do you think Baxter was?'', and ``To what extent do you feel frustrated by the instructions and behaviour of Baxter while working with it?''.
Additionally, we included a question to rate the extent to which the robot is {\em trustworthy}. 
The human nature, human uniqueness, Godspeed questionnaire items and trustworthiness were applied considering the participants' perceptions before and after the experiment took place, to measure a possible change in perception. 

Validated questions were used when possible, such as the TIPI, NARS, and the Godspeed questionnaire items. We followed the procedures reported in the literature to process the data. 
Furthermore, we conducted a reliability analysis for the customized scales through the Cronbach's $\alpha$ statistic, as reported in Table~\ref{t:alphas}.
An $\alpha>0.7$ or higher is considered acceptable, indicating reliability of the measuring scales, which is the result for most of the ones used in this article. 
This procedure is recommended as good practice by~\cite{Bartneck2009}, before ``calculating the mean scores'' for any scale.

\begin{table}
\caption{Cronbach's $\alpha$ statistic for customized scales\label{t:alphas}}
\centering
\begin{tabular}{|l|l|}
\hline
SCALE   & $\alpha$ \\\hline
Technological enthusiasm & 0.928\\ \hline
Human nature    & 0.428\\\hline
Human uniqueness & 0.601\\\hline
Psychological closeness     & 0.794\\\hline
Competence & 0.782\\ \hline
\end{tabular}
\end{table}%

Five-point Likert scales were employed for most of the items in the questionnaires, with scores of 1 to 5 to indicate ``strongly disagree'', ``disagree'', ``neither agree nor disagree'', ``agree'', and ``strongly agree'', respectively.
For a small number of the questions about previous experience with technology, and perceived presence of faults in the robot's operation, a ``yes'' or ``no'' answer was requested.

Finally, {\em qualitative data} was gathered via interviews with the participants. 
The interviews sought to further establish how the participants felt during the experiment, and their emotional state before and after interacting with Baxter. 
Questions included: ``Did your expectations about the experiment and the robot change after interacting with Baxter?'', ``Did you feel you were helping Baxter or Baxter was helping you, or you though it was more a mutual collaboration?'', ``What kind of improvements, e.g.\ for friendliness, could be done to Baxter?'', ``What would make you trust Baxter more?''.

\section{RESULTS}\label{sc:results}

In this section, we present the results of the subjective and objective assessments of the human--robot interaction, set in the context of a cooperative manufacturing task. 
Underlying data are openly available online.\footnote{From DOI: 10.5523/bris. ... }

We first applied the Saphiro-Wilk test to determine whether all the measured variables showed normal distributions. 
The Shapiro-Wilk test indicated the NARS data had normal distributions for the three subscales, along with the human nature, human uniqueness and psychological closeness.
However, the Godspeed questionnaire items (to assess animacy, likeability, intelligence, and safety), along with the scales for trustworthiness and competence, did not have a normal distribution in general.

For the resulting normal variables, parametric independent $t$-tests were used to compare pairs of groups of participants (e.g.\ exposed to a faulty robot or a robot with correct instructions, with and without previous experience). 
For the variables that did not show normal distributions, non-parametric Mann-Whitney $U$-tests were employed to compare pairs of groups, and Kruskal-Wallis tests for more than 3 groups (e.g.\ types of faults). 
Correspondingly, paired $t$-tests (parametric) or Wilcoxon signed rank tests (non-parametric) were used to determine the differences in participants' perception of the robot (e.g.\ human uniqueness, human nature, likeability, trustworthiness) before and after the assembly task. 
Pearson's (parametric) or Spearman's (non-parametric) correlation was used to analyse the effect of personality traits in the participants' subjective perception of the robot. 

Fisher's exact test was used to analyse the effect of condition, previous experience and personality (nominal) in the participants' performance when completing the race car. 

\subsection{Participants}

The participants were recruited by distributing posters in public notice boards in universities and libraries, and by word of mouth. 
We had 20 participants in total, but discarded the data of two of them due to malfunction of the video recording. 
From the 18 considered participants, 4 were female, 15 were male, and 1 preferred not to disclose the gender. 
Their ages ranged from 18 to 49 years old, with 12 participants between 18 and 24 years old, 4 between 25 and 29 years old, and 2 between 30 and 39 years old. 

Most of the participants (83.3\% ) were professionally involved in a science, technology, engineering and mathematics (STEM) subject --including 33.3\% of the total in the robotics field--, and 72.2\% admitted they had interacted with robots before.
The majority answered that they use computers frequently in their work/studies (mean $M=4.722$, standard deviation $SD=0.752$).
Also, the majority of the participants (72.2\% ) admitted they would be comfortable interacting with robots.

\subsubsection{Personality Traits}

The results on the personality test are reported in Table~\ref{t:personality}, from the indicated mean based analysis procedure in~\cite{Gosling2003}. 
In general, the participants were slightly extroverted and emotionally stable. 
Considering the mean of the scores (as the personality traits obeyed a normal distribution according to the Saphiro-Wilk test), 8 participants were extroverted (and 10 not extroverted), and 9 reported to be emotionally stable (and 9 not emotionally stable).

\begin{table}
\caption{Personality results for the group of participants, $N=18$\label{t:personality}}
\centering
\begin{tabular}{|c|l|l|l|l|l|l|}
\hline
 & \multicolumn{5}{|c|}{PERSONALITY CATEGORIES} \\ 
& E & A & C & ES & O\\ \hline
Mean ($M$)   & 3.556 & 3.833 & 3.639 & 3.806 & 3.583\\\hline
Standard Deviation ($SD$) 		& 0.802 & 0.822 & 0.854 & 0.770 & 0.895\\\hline
\end{tabular}

\footnotesize{Note:E, Extroversion; A, Agreeableness; C, Conscientiousness; ES, Emotional Stability; O, Openness.}
\end{table}%

\subsubsection{Technological Enthusiasm and Expectations on Robots}

In terms of technological enthusiasm, 13 out of 18 participants declared to be interested in technology and keep updated with the latest trends according to the mean of their resutls ($M=4.15$, $SD=1.055$). 
In terms of expectations for a robot, the participants as a group did not have polarized opinions with respect to a robot developing its own intelligence ($M=3.111$, $SD=1.324$), or the design of trustworthy robots ($M =3.222$, $SD = 1.166$). 
Approximately half of the participants (8 and 10, for intelligence and trustworthiness, respectively) had expectations above average.

\subsection{General Results}

The results of the NARS questions are shown in Table~\ref{t:nars}, grouped in the three different subscales, as mentioned in the previous section. 
The attitude of the group of participants towards robots and human-robot interactions is positive in general, as shown by the means according to the five-point Likert scale --i.e.\ the participants mostly disagreed or neither agreed or disagreed with having negative attitudes and anxieties towards robots and human-robot interactions. 
Considering the mean (as these items were found to obey a normal distribution by the Shapiro-Wilk test), 10, 6 and 10 participants had negative attitudes towards interactions with robots, the social influence of robots, and the emotions they felt when interacting with robots, respectively.

\begin{table}
\caption{NARS results for the group of participants, $N=18$\label{t:nars}}
\centering
\begin{tabular}{|cc|l|l|l|}
\hline
 && \multicolumn{3}{|c|}{NARS SUBSCALE} \\ 
&& S1 & S2 & S3\\ \hline
Mean ($M$)    && 2.367 & 2.744 & 2.708\\\hline
Standard Deviation ($SD$) 		&& 0.626 & 0.569 & 0.888\\\hline
\end{tabular}

\footnotesize{Note:S1, Negative attitude towards situations of interaction with robots; S2, Negative attitude towards social influence of robots; S3, Negative attitude towards emotions in interaction with robots. }
\end{table}%

\begin{figure}
\centering
\includegraphics[width=\textwidth]{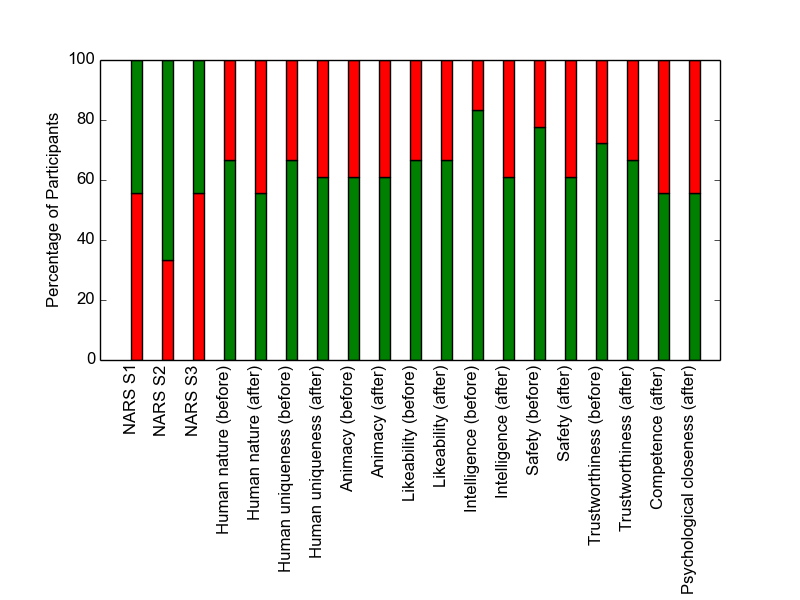}
\caption{Summary of results for the quantitative data analysis, with percentages of participants that rated their perception of the robot positively (green) or not (red).}
\label{fig:summaryraw}
\end{figure}

Approximately between 83.2\% (15 of 18) and 61.1\% (11 of 18) of the participants perceived anthropomorphic traits in the robot (above the mean for human uniqueness and human nature, or the median for likeability, animacy, intelligence, and safety), although these reduced to between 66.6\% (12 of 18) and 55.5\% (10 of 18) after the manufacturing task. 
Also, 72.2\% of the participants (14 of 18) found Baxter highly trustworthy (which reduced to 66.6\% after the interaction), but only 55.5\% found it highly competent overall (both above the median). 
Additionally, 55\% of the participants confirmed that they would be happy to work with Baxter again (psychological closeness), which indicates that participants found the collaborative task enjoyable despite the performance pressure enacted by timing constraints in the experiment (explained in the experimental setup in Section~\ref{ssc:experiment}). 
A summary of the results is shown in Figure~\ref{fig:summaryraw}.

During the experiment, 50\% of the participants struggled to assemble or disassemble the race car, needing assistance by the human operator. 

\subsection{Effect of Condition}\label{sc:condition}
\begin{figure}
\centering
\includegraphics[width=\textwidth]{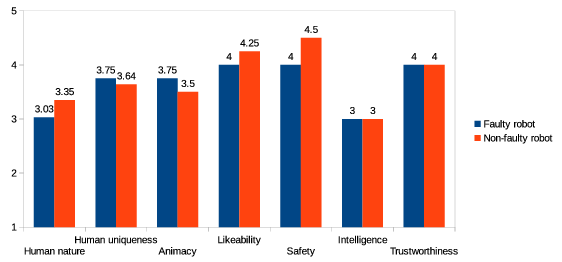}
\caption{Computed mean or median of perceived robot characteristics before the manufacturing task, according to the condition grouped as faulty or non-faulty. A score of 1 indicates low perception of a trait, whereas a score of 5 indicates a high perception of a trait.}
\label{fig:condition1}
\end{figure}

\begin{figure}
\centering
\includegraphics[width=\textwidth]{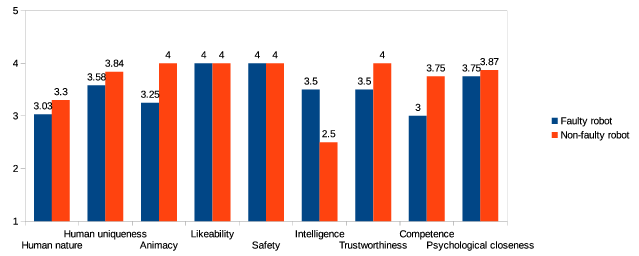}
\caption{Computed mean or median of perceived robot characteristics after the manufacturing task, according to the condition grouped as faulty or non-faulty. A score of 1 indicates low perception of a trait, whereas a score of 5 indicates a high perception of a trait..}
\label{fig:condition2}
\end{figure}

The participants were divided into two groups, the ones that experienced an interaction with a non-faulty robot, and the ones that interacted with a faulty robot. 
According to the recorded data, 5 of the participants perceived at least one fault in Baxter's behaviour, but actually collaborated with a non-faulty robot. 
On the contrary, a participant interacted with Baxter providing faulty assembly instructions, but reported the perception of no fault. 
A summary of the results, showing the mean or median (according to the Shapiro-Wilk test results), is shown in Figure~\ref{fig:condition1} before the manufacturing task, and in Figure~\ref{fig:condition2} after the manufacturing task. 

The condition (i.e.\ a faulty or non-faulty robot) did not have an effect on the participants' subjective perception of Baxter in terms of human nature ($t(df=16)=-2.024, p > 0.05$), human uniqueness ($t(df=16)=1.222, p > 0.05$), likeability ($U=35, p>0.05$), animacy ($U=34.5, p>0.05$), safety ($U=32, p>0.05$), and intelligence ($U=39.5,p>0.05$) before the manufacturing task. 
The same effect is observed after the manufacturing task ($t(df=16)=-1.061, p> 0.05$; $t(df=16)=-0.540, p > 0.05$; $U=32.5,p>0.05$; $U=30,p>0.05$; $U=37.5, p>0.05$; and $U=29.5,p>0.05$, for the items in the same order as before). 
Likewise, no significant effect was found in the reported perceived psychological closeness to the robot ($t(df=16)=-0.270,p>0.05$), or its perceived competence ($U=27.5,p>0.05$). 
The perceived trustworthiness did not differ as a cause of the condition, either before ($U=28.5,p>0.05$) or after ($U=20.5,p>0.05$) the task.

We also evaluated the effect of the type of fault ($A$, $B$ or $C$ from Table~\ref{t:conditions}) in the participants' perception of the robot for some of the traits, with no statistically significant results, for intelligence ($\chi^2(df=2)=0.638, p>0.05$ before the task; $\chi^2(df=2)=0.132, p>0.05$ after the task), safety ($\chi^2(df=2)=1.750, p>0.05$ before the task; $\chi^2(df=2)=1.481, p>0.05$), trustworthiness ($\chi^2(df=2)=5.500,p>0.05$ before the task; $\chi^2(df=2)=0.132, p>0.05$ after the task), and competence ($\chi^2(df=2)=1.187, p>0.05$).

Comparing the participant's perception of the robot before and after the cooperative manufacturing task, results indicate no statistically significant change in perceived human nature and human uniqueness for the participants in the faulty condition ($t(df=7)=0.032, p>0.05$ and $t(df=7)=1.384, p>0.05$) in any of the three fault variants together, and in the non-faulty condition ($t(df=9)=0.271, p > 0.05$ and $t(df=9)=-0.054, p>0.05$). 
In the results for the Godspeed questionnaire scales, perceived safety ($Z=-1.414,p>0.05$) and intelligence ($Z=-1.300,p>0.05$) decreased for the participants with the non-faulty condition, whereas animacy ($Z=-2.070,p<0.05$) increased.
Perceived safety ($Z=-1.342,p>0.05$) decreased for the participants that interacted with the faulty robot. 
Perceived likeability did not present a statistically significant change regarding the non-faulty ($Z=-0.343,p>0.05$) or faulty  ($Z=-0.333,p>0.05$) conditions, nor perceived animacy ($Z=-0.957,p>0.05$) and intelligence ($Z=0.000,p>0.05$) for the faulty condition. 
Perceived trustworthiness was found to increase for the non-faulty condition ($Z=-1.414, p>0.05$), whereas it decreased for the faulty condition in general ($Z=-0.557,p>0.05$). 
A summary of the found changes before and after the task is presented in Table~\ref{t:changetraits}.

\begin{table}
\caption{Summary of results for the effect of condition in the change in perceived traits\label{t:changetraits}}
\begin{tabular}{|c|l|l|}
\hline
SCALE/MEASURE & NON-FAULTY CONDITION & FAULTY CONDITION \\ \hline
Human nature change   & No change & No change\\\hline
Human uniqueness change		& No change & No change\\\hline
Animacy change & Increased & No change \\ \hline
Likeability change & No change & No change \\ \hline
Safety change & Decreased & Decreased \\ \hline
Intelligence change & Decreased & No change \\ \hline
Trustworthiness change &  Increased & Decreased \\ \hline
\end{tabular}
\end{table}%

In the objective assessment of the experiment, the condition did not have a statistically significant effect on the participant's completion of the manufacturing task ($\chi^2(df=1)=3.6, p>0.05$), i.e.\ without struggling to follow the instructions and thus receiving help by one of the operators.

\subsection{Effect of Previous Experience}\label{sc:experience}

The participants were divided into four groups: the participants that had previous experience, and the ones without previous experience with robots; and the enthusiastic ones about technology, and the ones that were not so enthusiastic (according to the mean of their score). 
We analysed the effect of previous experience with robots and participants' interest in technology on negative attitudes towards robots and technology, and the subjectively assessed robot's traits. 

The negative attitudes towards robots and technology do not present a statistically significant difference for the participants with and without previous experience with robots, for the three NARS subscales ($t(df=16)=0.690,p>0.05$; $t(df=16)=-1.625, p>0.05$; and $t(df=16)=-0.556, p>0.05$, respectively for the scales S1, S2 and S3). 
Similar results were found for participants with and without technological enthusiasm ($t(df=16)=-1.349,p>0.05$; $t(df=16)=-0.431, p>0.05$; and $t(df=16)=-1.173, p>0.05$).

\begin{figure}
\centering
\includegraphics[width=\textwidth]{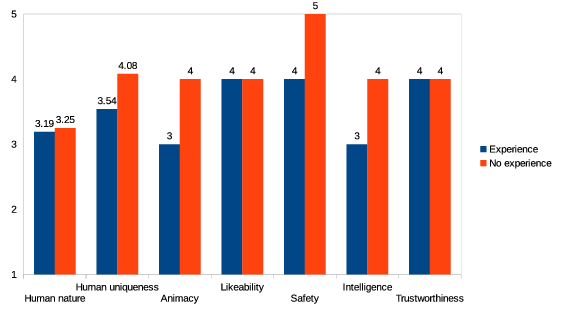}
\caption{Computed mean or median of perceived robot characteristics before the manufacturing task, according to previous experience with robots. A score of 1 indicates low perception of a trait, whereas a score of 5 indicates a high perception of a trait.}
\label{fig:exp1}
\end{figure}

\begin{figure}
\centering
\includegraphics[width=\textwidth]{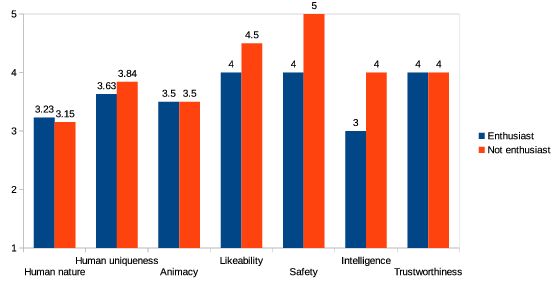}
\caption{Computed mean or median of perceived robot characteristics before the manufacturing task, according to enthusiasm about technology. A score of 1 indicates low perception of a trait, whereas a score of 5 indicates a high perception of a trait.}
\label{fig:enth1}
\end{figure}

Before the cooperative manufacturing task, previous experience with robots did not show a statistically significant effect on the participants' perception of the robot regarding human nature ($t(df=16)=-0.161, p>0.05$) or human uniqueness ($t(df=16)=-0.795, p>0.05$). 
The same lack of effect was found for participants' technological enthusiasm with respect to perceived human nature ($t(df=16)=0.226, p>0.05$) and human uniqueness ($t(df=16)=-0.686, p>0.05$). 
Also, no statistically significant effect was found for previous experience with robots and technological enthusiasm, with respect to perceived likeability ($U=28.5, p>0.05$; $U=24, p>0.05$), animacy ($U=17, p>0.05$; $U=32.5, p>0.05$), safety ($U=19, p>0.05$; $U=24, p>0.05$), intelligence ($U=20, p>0.05$; $U=24.5, p>0.05$), and trustworthiness ($U=20.5, p>0.05$; $U=25, p>0.05$). 
A summary of the results is shown in Figure~\ref{fig:exp1} considering previous experience with robots, and in Figure~\ref{fig:enth1} considering enthusiasm about technological advances. 

After the cooperative manufacturing task, previous experience with robots also did not show a statistically significant effect on the participants' perception of the robot regarding human nature ($t(df=16)=-1.282, p>0.05$) or human uniqueness ($t(df=16)=-1.476,p>0.05$). 
Similar results were found for participants' technological enthusiasm with respect to human nature ($t(df=16)=0.846, p>0.05$) and human uniqueness ($t(df=16)=-0.795, p>0.05$). 

\begin{figure}
\centering
\includegraphics[width=\textwidth]{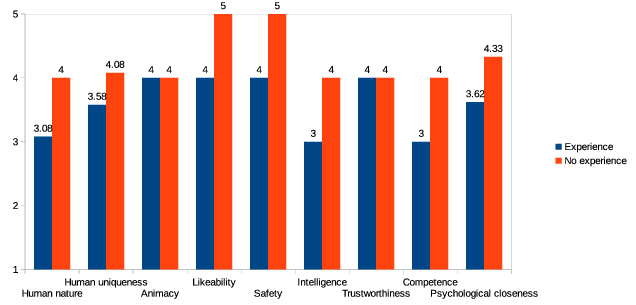}
\caption{Computed mean or median of perceived robot characteristics after the manufacturing task, according to previous experience with robots. A score of 1 indicates low perception of a trait, whereas a score of 5 indicates a high perception of a trait.}
\label{fig:exp2}
\end{figure}

\begin{figure}
\centering
\includegraphics[width=\textwidth]{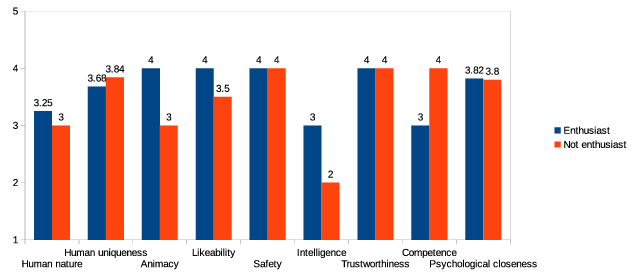}
\caption{Computed mean or median of perceived robot characteristics after the manufacturing task, according to enthusiasm about technology. A score of 1 indicates low perception of a trait, whereas a score of 5 indicates a high perception of a trait.}
\label{fig:enth2}
\end{figure}

With respect to the Godspeed questionnaire scales, previous experience with robots and technological enthusiasm did not have a statistically significant effect on perceived animacy ($U=27, p>0.05$; $U=27.5, p>0.05$), likeability ($U=16, p>0.05$; $U=27.5,p>0.05$), safety ($U=18, p>0.05$; $U=26, p>0.05$), and trustworthiness ($U=15.5, p>0.05$; $U=26.5, p>0.05$). 
Previous experience with robots did have an effect on perceived intelligence ($U=12, p<0.05$), with the participants that had never interacted with robots before giving higher scores to Baxter. 
Technological enthusiasm did not have an effect on perceived intelligence ($U=32,p>0.05$). 
A summary of these results is shown in Figure~\ref{fig:exp2} considering previous experience with robots, and in Figure~\ref{fig:enth2} considering enthusiasm about technological advances.

Neither previous experience with the robot nor enthusiasm about technology had a statistically significant effect on perceived psychological closeness to the robot, as reported by the participants ($t(df=16)=-1.608, p>0.05$; $t(df=16)=0.043, p>0.05$).
The same was observed for the perceived robot's competence ($U=18.5, p>0.05$; $U=22 p>0.05$).

Previous experience with robots or enthusiasm about technology did not have a statistically significant effect on the completion of the task without help ($\chi^2(df=1)=0.277, p>0.05$; and $\chi^2(df=1)=2.492, p>0.05$ respectively).

\subsection{Effect of Participant's Personality}\label{sc:personality}

As reported in~\cite{Salem2015}, we analysed the effect of participants' personality traits, in particular extroversion and emotional stability, in their perception of the robot. 
Extroversion correlated negatively with the NARS scales S1 and S3, i.e.\ negative perception of interactions with robots and emotions during the interactions ($r_s=-0.511, p<0.05$ for S1; and $r_s=-0.791, p<0.01$ for S3, respectively). 
This means that extroverted participants perceive interactions with robots more positively overall. 
In the other scale S2 (negative perception of social impact of robots and technology) no statistically significant correlation was found ($r_s=-0.031, p>0.05$).
The other trait, emotional stability, did not show a statistically significant correlation with the NARS subscales ($r_s=-0.332, p>0.05$ for S1; $r_s=-0.214, p>0.05$ for S2; and $r_s=-0.411, p>0.05$ for S3).

Before the cooperative manufacturing task took place, extroversion negatively correlated with the participants' perception of the robot's human nature ($r_s=-0.763, p<0.01$). 
No statistically significant correlation was found between extroversion and the participants' perception of the robot's human nature after the manufacturing task ($r_s=0.217, p>0.05$), nor for other robot traits before and after the task, including human uniqueness ($r_s=0.130, p>0.05$ for before the task; $r_s=0.208, p>0.05$ for after the task), animacy ($r_s=0.034, p>0.05$ before; $r_s=0.072, p>0.05$ after), likeability ($r_s=-0.053, p>0.05$ before; $r_s=0.231, p>0.05$ after), intelligence ($r_s=0.040, p>0.05$ before; $r_s=-0.121, p>0.05$ after), safety ($r_s=0.105, p>0.05$ before; $r_s=0.121, p>0.05$ after), and trustworthiness ($r_s=0.123, p>0.05$ before; $r_s=0.257, p>0.05$ after). 
Also, extroversion did not correlate to the perception of the robot's competence $r_s=-0.369, p>0.05$, nor with psychological closeness between the participants and the robot ($r_s=0.181,p>0.05$).

Emotional stability was found to positively correlate to perceived psychological closeness ($r_s=0.518, p<0.05$), although no statistically significant correlation was found with the perception of the robot's competence ($r_s=-0.071, p>0.05$). 
Emotional stability did not show a statistically significant correlation, for both before and after the interaction respectively, with perceived human nature ($r_s=-0.427, p>0.05$; $r_s=0.324, p>0.05$), human uniqueness ($r_s=.417, p>0.05$; $r_s=0.373, p>0.05$), animacy ($r_s=0.072, p>0.05$; $r_s=0.412, p>0.05$), likeability ($r_s=0.192, p>0.05$; $r_s=0.372, p>0.05$), intelligence ($r_s=0.390, p>0.05$; $r_s=-0.044, p>0.05$), safety ($r_s=0.072, p>0.05$; $r_s=0.296, p>0.05$), and trustworthiness ($r_s=0.085, p>0.05$; $r_s=.281, p>0.05$).

Extroversion and emotional stability did not have a statistically significant effect on the completion of the manufacturing task without help ($\chi^2(df=2)=1.111, p>0.05$; and $\chi^2(df=1)=0.222, p>0.05$, respectively).

\subsection{Effect of human-robot interaction task}\label{sc:hri}

We compared our results with those reported in~\cite{Salem2015,Salem2015challenges}, in the context of a home care assistant with a wide range of `faulty' behaviours, from taking detours when navigating towards a location, following unusual requests such as using somebody else's password or pouring orange juice into a plant pot. 
In terms of the condition, they reported that the non-faulty robot was found to be more trustworthy, competent, and uniquely human.
In contrast, in our collaborative manufacturing scenario, we did not find any statistically significant effect of the implemented instruction faults in the participants' perception of the robot overall.

In terms of correlation between the participants' personality traits and the participants' perception of the robot,~\cite{Salem2015} found a positive correlation between extroverted participants and their perception of human nature, human uniqueness, and psychological closeness with the robot. Alongside, they found a positive correlation between emotional stability and perceived animacy, likeability and psychological closeness. 
In our scenario, we found that extroversion was negatively correlated with the perception of human nature, contrary to~\cite{Salem2015}, and emotional stability was positively correlated with perceived psychological closeness, in agreement with~\cite{Salem2015}. 
Consequently, we conclude that our results for the subjective assessment of effects of condition and personality conflict with the ones in~\cite{Salem2015}. 

On the other hand, the objective assessment results agree with~\cite{Salem2015}, as personality did not have an effect on the performance of participants in the human-robot interactions. 
Other elements and results in their study cannot be compared to ours.

\subsection{Qualitative Data Analysis}\label{sc:qualitative}

We classified the answers to open-ended questions in the post-experiment interviews, based on the collected data. 
For the question ``Did your expectations about the experiment and the robot change after interacting with Baxter?'', the categories were (ordered from high to low incidence within participants): 
\begin{itemize}
\item {\em Low expectations, but pleasantly surprised}. (61\% of participants)\\
\item {\em Motion and intelligence did not meet expectations}. (22.2\% of participants)
\item {\em Speech and interfaces did not meet expectations}. (11.1\% of participants)
\item {\em Manufacturing task and instructions did not meet expectations}. (11.1\% of participants)
\item {\em Expectations were met}. (5.5\% of participants)
\end{itemize}

The answers to the question ``Did you feel you were helping Baxter or Baxter was helping you, or you thought it was more a mutual collaboration?'' were divided into: {\em the participant helped the robot} (27.8\% of participants); {\em the robot helped the participant except when the robot made mistakes} (11.1\% of participants); and {\em collaboration between the participant and the robot} (22.2\% of participants).

The answers to the question ``What would make you trust Baxter more?'' were classified into (ordered from high to low incidence within participants):
\begin{itemize}
\item {\em Participant teaching or programming the robot}. (22.2\% of participants)
\item {\em Demonstration and proof}. (22.2\% of participants)
\item {\em A competent programmer}. (16.6\% of participants)
\item {\em A simple task where risks are minimal}. (11.1\% of participants)
\item {\em A robot with artificial intelligence}. (5.5\% of participants)

\end{itemize}

Finally, the answers to the question ``What kind of improvements, e.g.\ for `friendliness', could be done to Baxter?'' were divided into the categories (ordered from high to low incidence within participants):
\begin{itemize}
\item {\em Better interfaces}, e.g.\ voice, face, body language (66.7\% of participants). 
\item {\em Feedback on the task}, e.g.\ acknowledgement of participant's actions (11.1\% of participants). 
\item {\em Added artificial intelligence}, for smoother motion, to avoid mistakes (5.5\% of participants).
\item {\em Better motion}, e.g.\ human-like (5.5\% of participants). 
\end{itemize}

\section{DISCUSSION}\label{sc:discussion}

In this section we analyse whether the reported results support or disprove the hypotheses listed in Section~\ref{sc:hypotheses}, and comment on threats to the validity for the experiment and results. 

\subsection{Hypotheses}

Although the results regarding differences in the participants' perception of the robot, for those who interacted with a faulty robot or those who did not in Section~\ref{sc:condition}, do not support {\em Hypothesis 1 (a)}, the changes in their perception before (when meeting the robot) and after the interaction do support an effect on perceived animacy, safety, intelligence and trustworthiness as a consequence of the condition.
Also, contrary to our intuition, perceived safety and intelligence decreased for the participants that interacted with the non-faulty robot, which suggests that the robot's design and behaviour did not meet their expectations. 
Additionally, the type of manipulation (minor fault to severe fault) did not show a statistically significant effect on the participants' perception of the robot.
Consequently, our results contradict the ones in~\cite{Salem2013,Salem2015}, as the robot is perceived in a similar manner regardless whether it is faulty or not.
These differences could be caused by the type faults (i.e.\ cognitive) that we implemented in our scenario, such as giving the wrong instruction to the participant, which do not impede the possibility of a successful manufacturing outcome.
Physical faults such as breaking one of the pieces have irreversible and dangerous consequences, and thus could have a stronger effect in participants' perceptions compared to cognitive faults. 
Another cause could be the nature of the task in our experiment, i.e.\ an industrial or `working environment' context compared to a more `social' setting in~\cite{Salem2015}, were the participants do not connect with the robot and thus fail to perceive the faults.
Alternatively, the differences could be caused by the difficulty of the task, as some of the participants commented ``it was hard to follow'', and thus many of the participants perceived the robot as ``clumsy'' regardless of the presence of faults.

The results in Section~\ref{sc:experience} support {\em Hypothesis 2 (a)}, which predicted an effect on participants' perception due to previous experience with robots and enthusiasm about technology, as participants that had never interacted with robots before gave higher intelligence scores to Baxter after the interaction took place. 
This is reflected by the qualitative data in Section~\ref{sc:qualitative}, where many of the participants stated that the expectations they had on the robot's capabilities were surpassed, specially if they had not interacted with robots before. 
No other effect of experience and technological enthusiasm was found on the measured participants' perception of the robot.  
The effect of pre-experiment training with Baxter for participants that had never interacted with robots before needs to be further investigated, as in real life collaborating with a robot in the workspace would involve a familiarization and training stage.

The results in Section~\ref{sc:personality} support an effect of participants' personality in their perception of the robot according to {\em Hypothesis 3 (a)}, as correlations were found between extroversion and perceived human nature, and emotional stability and psychological closeness with the robot.
Nonetheless, we expected extroverted and emotionally stable participants to perceive Baxter more human-like -- although no effect was expected in trustworthiness--, as reported in~\cite{Salem2015}. 
Our results suggest, again, that the context of the task in a `working environment', along with the robot and the experiment's design, might have an effect on perceived anthropomorphism, competence and intelligence.

The results of the objective assessment invalidate {\em Hypotheses 1 (b), 2 (b)} and {\em 3 (b)}, as the presence of faults did not have an effect on the successful completion of the manufacturing task (without receiving help from one of the operators), and the same results were found for experience with robots and technological enthusiasm, and personality traits such as extroversion and emotional stability.

Finally, the results in Section~\ref{sc:hri} support {\em Hypothesis 4}, as the overall results differ according to the type of human-robot interaction.
In a home care and more social setting, faults in the robot and the participants' personality affected the perception of the robot as human-like, competent and trustworthy, in a more noticeable manner. In contrast, there were fewer statistically significant effects in our manufacturing setting. 
These comparisons suggest that, as the participants were following a fast sequence of displayed instructions, without previous training, feedback from the robot, and multiple interaction interfaces such as voice, the subtlety of faults might have minimized their own effect on the participant's perception of the robot's competence. 
On the other hand, the qualitative data also suggests that an ``interactive'' and human-like robot is rarely associated with an industrial high-precision automated arm, for which high-standards of safety and reliability are expected, implemented by a competent programmer. 
Some of the faults programmed in Baxter's instructions were perhaps perceived as `part of the intelligence' of the robot, acknowledging it did something wrong, but capable of correcting it. 
A more comprehensive study of humanoid and interactive assistants in industrial manufacturing settings is required to understand the formation of trust, compared to other collaborative work scenarios such as in~\cite{Desai2012,Desai2013}, and more social ones such as the ones in~\cite{Salem2015,Salem2015challenges}.

In our manufacturing scenario, the robot is fixed at one location and `works' with the participant in a manufacturing cell that contains other industrial robots around.
Consequently, the participants might not see the robot as a `friend' but more as a `friendly industrial tool'. 
In contrast, in~\cite{Salem2015} the robot inhabits a home and the actions are about social and homely activities such as tidying up, or walking around the house.  
The physical design of the robot adds another research dimension to the interactions, as robots such as the Care-O-bot used in~\cite{Salem2015} have voice and a face to interact with the participants, whereas robots like Baxter might only have a screen with a displayed face.

\subsection{Critical Analysis of Validity of Findings}

\subsubsection{Experiment} 
Possible threats to internal validity in the design of the experiment include the introduced faults in the sequences of instructions, which might have been subtle to the participant, and not as threatening in terms of consequences as the ones used in~\cite{Salem2015}; and the level of difficulty of the task, which might have biased the participants' perception of the robot's capabilities and human-like traits related to friendliness and animacy, and reduced psychological closeness. 
The design of the robot's communication interfaces might have also played a role in biasing the participants' perception of the robot, as other experiments have used combinations of visual, gestures and sound (i.e.\ voice) channels to provide feedback besides guidance, as e.g.\ in~\cite{Salem2011,Hamacher2016}. 
Although we considered adding artificial intelligence to provide feedback on the participant's correctness of their actions, or to achieve a greater precision in manipulation, both then become variables to evaluate when investigating whether trust and confidence in the robot increase, perhaps at the cost of psychological closeness and perceived human-like traits.

Threats to external validity include environmental factors that might have influenced the results such as the location of the experiment, in a robotics laboratory, and an evident presence of the operators in the background. 
Also, participants had a limited range of age, and many of them had occupations related to STEM subjects and thus close contact with technology, which might have influenced the results overall.

A larger set of participants with a wider age range and variety of experience with technology is needed to further validate the results of this experiment.
More importantly, interactions between potential users, e.g.\ current manufacturing workers, and industrial robotic assistants need to be investigated. 
In addition, experiments with exposition to longer interactions with the robot are needed, including ones that contain some social interaction as well, equivalent to chatting with colleagues during a break from work. 
Training or a familiarization phase with the robot before the actual experiments could help to provide a more detailed picture on possible improvements towards increasing people's trust in the robot, once the users have understood what the functional capabilities of such a robot are.

\subsubsection{Measured Variables}

Validated questions were used when available for both the pre- and post-questionnaires.
Nonetheless, some items such as trustworthiness were evaluated only through single scale measurements, which might be less reliable than combining several, but minimal scales such as it is done for the TIPI or NARS. 
Note that we did not measure reliability, coactivity, productivity and efficiency~\cite{Murphy2013}, which are alternative popular metrics to assess human-robot interactions, as we were only interested in evaluating trust and not assessing the robot's design. 
Although more generalizable and human-robot interaction specific metrics are needed to be able to compare a particular scenario or application with another~\cite{Murphy2013}.

\section{CONCLUSION}\label{sc:conclusion}

We presented the results of an experiment with a real robot and participants executing a collaborative manufacturing task, where a race car was assembled first and then disassembled, in order to study factors that influence the development of trust in this context. 
We designed the experiment to observe the impact of faults of different kinds during the manufacturing process, as well as participants' previous experience and personality traits, in trust measured as perceived competence, safety, and trustworthiness, and the performance of the participants (both objective and subjective assessments).

The results show that, compared to other scenarios such as in~\cite{Salem2015,Salem2015challenges}, the presence of faults did not have a significant effect on the participants' perception of the robot or their performance in the task. 
Nonetheless, participants' experience and personality did have an effect on their perception of the robot, although relatively small when compared to the one reported in~\cite{Salem2015}. 
Some of the surprising results could be associated with the nature of the task (difficult and demanding), the subtlety of the errors in the robot's functioning, and the lack of training (which would be expected in real-life manufacturing work). 
Consequently, longer studies with a wider variety of faults, and possibly pre-training in the task, are needed to further understand the development of trust in collaborative manufacturing and to validate the results of our study. 
Understanding in deeper detail what makes robotic assistants trustworthy will facilitate collaboration with robots in shared workplaces, creating effective and satisfying working environments.

\subsection*{Acknowledgements}
We would like to thank Carl and Lena Richards for donating their toy kit to this research project.

The work by D. Araiza-Illan and K. Eder was partially supported by the Engineering and Physical Sciences Research Council (EPSRC), grants EP/J01205X/1 RIVERAS: Robust Integrated Verification of Autonomous Systems and EP/K006320/1 Trustworthy Robotic Assistants.

\end{document}